\documentclass[runningheads]{llncs}
\usepackage[T1]{fontenc}
\usepackage{graphicx}
\usepackage{hyperref}
\usepackage{color}

\usepackage{amssymb}
\usepackage{pifont}

\usepackage{subcaption}

\usepackage{amsmath}

\usepackage{multirow}
\usepackage{adjustbox}
\usepackage{tikz}
\usepackage{xcolor}

\definecolor{goldenrod}{RGB}{218, 165, 32}
\definecolor{royalblue}{RGB}{65, 105, 225}

\definecolor{darkGreen}{rgb}{0.0, 0.5, 0.0}

\newcommand{\greenTriangleReverse}{\tikz \fill [goldenrod] (0,0) -- ++(0:1.8ex) -- ++(-120:1.8ex) -- ++(-240:1.8ex) -- cycle; \hspace{0.2em}}
\newcommand{\blueTriangleReverse}{\tikz \fill [royalblue] (0,0) -- ++(0:1.8ex) -- ++(-120:1.8ex) -- ++(-240:1.8ex) -- cycle; \hspace{0.2em}}
\newcommand{\blueSquare}{
    \tikz \fill [royalblue] (0,0) rectangle (1.8ex,1.8ex);
    \hspace{0.2em}
}

\newcommand{\greenSquare}{
    \tikz \fill [goldenrod] (0,0) rectangle (1.8ex,1.8ex);
    \hspace{0.2em}
}

\newtheorem{axiom}{Axiom}

\begin{document}
\title{Unsupervised Surrogate Anomaly Detection}

%
%
\author{Simon Klüttermann\orcidID{0000-0001-9698-4339} \and
Tim Katzke\orcidID{0009-0000-0154-7735} \and
Emmanuel Müller\orcidID{0000-0002-5409-6875}}

%
%
\institute{Department of Computer Science \\
TU Dortmund University \\
Dortmund, Germany\\
\email{Simon.Kluettermann@cs.tu-dortmund.de}}
%

\maketitle              
\begin{abstract}
In this paper, we study unsupervised anomaly detection algorithms that learn a neural network representation, i.e. regular patterns of normal data, which anomalies are deviating from. Inspired by a similar concept in engineering, we refer to our methodology as surrogate anomaly detection.
We formalize the concept of surrogate anomaly detection into a set of axioms required for optimal surrogate models and propose a new algorithm, named DEAN (Deep Ensemble ANomaly detection), designed to fulfill these criteria.
We evaluate DEAN on 121 benchmark datasets, demonstrating its competitive performance against 19 existing methods, as well as the scalability and reliability of our method.

\keywords{Anomaly Detection  \and Ensemble Methods.}
\end{abstract}

\section{Introduction}

Anomaly detection (AD) is a crucial subdomain of data analytics with countless applications ranging from fraud detection~\cite{fraudapl} to health monitoring~\cite{auto_appl_falldetection}. In all of these domains, anomalies are rare, exceptional or interesting objects that are highly deviating from the residual (regular) data~\cite{metasurvey}.
Generally, anomaly detection can be approached in three primary ways: with extensive access to labeled anomalies (supervised), with access to a limited number of labeled anomalies (semi-supervised), or without labeled anomalies (unsupervised). In this paper, we focus on unsupervised anomaly detection. This setting poses the unique challenge of identifying a wide range of anomalies without any predefined anomalous patterns or examples to follow.
At the same time, this approach is particularly valuable in real-world scenarios, where obtaining labeled data may be costly or impractical. Consequently, employing methods capable of detecting deviations from a purely data-driven inferred notion of normality becomes essential.

\begin{figure}[h]
  \centering
  \includegraphics[width=0.8\linewidth]{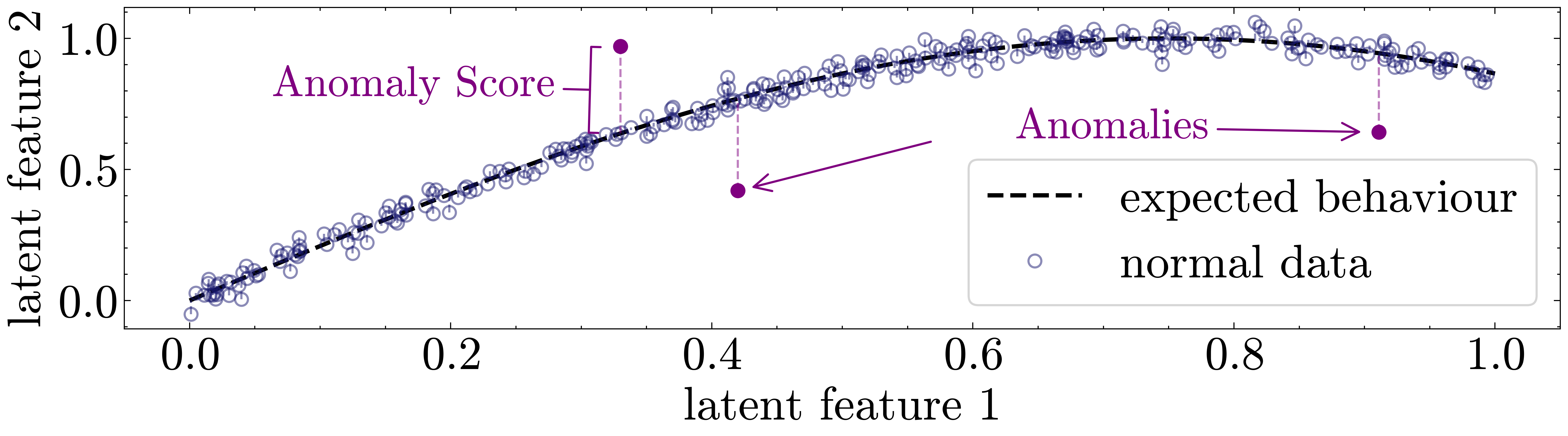}
  \caption{Example of surrogate anomaly detection: Anomalies are detected by learning a representation that encodes the regular patterns of normal data and measuring deviations from the expected behavior.}
  \label{fig:eye}
\end{figure}

To solve the task of unsupervised anomaly detection, various anomaly detection algorithms have been proposed. Methods such as AnoGan~\cite{gan} aim to model the probability density distribution of normal samples and consider samples in low-density regions as abnormal. Other algorithms, like KNN~\cite{otherknn}, consider samples that are further away in the variable space from other samples as more anomalous. Approaches like Isolation Forests~\cite{ifor} measure how easily samples can be separated. Still, it can be shown that both of these alternative approaches effectively model densities too~\cite{iforisdensity,knn}.

While density estimation is an intuitive solution to anomaly detection, any method based on this approach has two fundamental drawbacks. First, they do not scale well to high-dimensional data, which is known as the curse of dimensionality~\cite{curseOdim}. Second, there is usually no explicit modeling of the data-generating process involved, which limits how well the method generalizes to new data.

In contrast to modeling complex density distributions for high dimensional input data, we consider in this paper algorithms that learn low dimensional representations as approximations of the underlying regular patterns in the data. More specifically, we are inspired by surrogate models in engineering applications~\cite{surrogatesSummary}, where simple surrogate models are used to approximate more complex or expensive processes. Similarly, we search for models that capture a pattern of the underlying data-generating process instead of modeling the whole distribution, which we will subsequently refer to as surrogate anomaly detection models. An illustrative toy example of such a model is shown in Figure~\ref{fig:eye}. 

Some existing algorithms can be considered instantiations of such surrogate anomaly detection models. Autoencoder~\cite{aean} and PCA-based anomaly detection~\cite{pca} learns an identity function to compress regular data and measure the deviation of anomalies as the reconstruction error. DeepSVDD~\cite{deepsvdd} learns a representation in which regular data can be modeled by mapping it to a lower dimensional constant, where anomalies deviate highly from this constant value. Because these algorithms don't need to model the entire density distribution in its original input space, they scale more effectively to high dimensional data~\cite{aegoodhighdim}.


However, these methods are not without limitations either. Unlike density estimation methods, the objective of training a surrogate is much less well-defined, leading to an unlimited amount of options on how to create such a surrogate, each with its own drawbacks. We exploit this variability to formalize the idea of surrogate models into a blueprint for creating arbitrary surrogates. Within this framework, we propose five axioms that an ideal surrogate model should satisfy.
Based on these, we suggest a new surrogate AD algorithm called DEAN, which, to the best of our knowledge, is the first algorithm adhering to all of them.

We evaluate our algorithm by following the procedure outlined in a recent benchmark survey paper~\cite{surveyzhao}, comparing it against 19 competitors
across 121 datasets. Our algorithm performs highly competitively, showing only minor performance differences with the best non-surrogate competitors and outperforming all other surrogate-based methods.

Our main contributions are: (1) the formulation of a general framework for surrogate anomaly detection; (2) the establishment of guiding axioms for designing optimal surrogate algorithms; and (3) the development and comprehensive evaluation of a novel algorithm based on these principles.

To ensure the reproducibility of our results, our implementation is publicly available at \href{https://github.com/psorus/DEAN-model}{https://github.com/psorus/DEAN-model}.



\section{Related Work}

This section reviews related work with a focus on three key aspects: unsupervised anomaly detection, emphasizing approaches that extract meaningful patterns, ensemble methods, which, as discussed later, may enable the extraction of diverse patterns, and surrogate models in a more general context.

\subsection{Unsupervised Anomaly Detection}

Anomaly detection in an unsupervised setting inherently faces the challenge of defining a suitable objective without ground truth labels.
A common suggestion is to model the densities of normal, expected samples~\cite{metasurvey}, under the assumption that samples in low-density regions are less likely to be generated by the same process as normal data, and are therefore more likely to be anomalies. However, density estimation fundamentally suffers from the so called curse of dimensionality~\cite{curseOdim}, which limits how well these algorithms work on high-dimensional data.

Instead of modeling the densities of normal samples, certain anomaly detection methods extract a characteristic pattern that normal samples typically satisfy and test, whether new samples conform to this pattern. Examples of this are DeepSVDD~\cite{deepsvdd}, which tries to learn a representation in which every sample is mapped close to a certain point, or reconstruction-based methods like Autoencoder~\cite{aean} and PCA~\cite{pca}, which try to learn a lower dimensional latent representation, that captures all necessary information to reconstruct (only) normal samples.

These anomaly detection algorithms are less affected by the curse of dimensionality, because latent patterns typically do not increase significantly in complexity with additional features~\cite{aegoodhighdim}.
Still, these algorithms also have flaws. DeepSVDD requires careful training or will simply not perform well~\cite{surveyzhao}, and using reconstruction-based algorithms requires choosing a suitable size of the latent space, which is difficult without feedback through labeled anomalies~\cite{hypersomething}.
Thus, in this paper, we generalize such models and suggest an optimal approach based on axioms that outline essential properties.

\subsection{Ensemble Methods}

Ensembles are a powerful method in machine learning~\cite{aggarwalTheoryEnsembles}; techniques such as bagging, boosting, and stacking are well-established for combining multiple submodels into a superior model. In supervised tasks, the availability of labels facilitates the coordination of submodels, but in unsupervised settings, the absence of labels complicates this process. Thus, many unsupervised anomaly detection approaches simply aggregate the anomaly scores produced by various algorithms~\cite{plaincombine}. This strategy takes advantage of the fact, that errors made by diverse submodels tend not to be repeated across the entire ensemble~\cite{dean}.

One effective approach is the use of homogeneous ensembles, which merge many similar and simple submodels that, although weak individually, collectively yield robust performance through a combination of specialization and diversity~\cite{ifor,randnet}. Moreover, model-independent methods such as feature bagging~\cite{feature-bagging} can further enhance diversity by ensuring that submodels specialize in different subsets of data dimensions, which can also improve explainability~\cite{deanshap}.

\subsection{Surrogate Models}

Surrogate models are simplified approximations of more complex or computationally expensive models~\cite{surrogatesSummary}. In engineering, they are commonly employed when, for example, when physical simulations are too costly~\cite{surrogateWithML}. In machine learning, surrogates have been used to approximate, accelerate, or explain other machine learning models. This has been applied to uncertainty estimation~\cite{surrogateUQ}, explainability (both globally~\cite{globalsurrogates}, and locally~\cite{lime}), surrogate task-based models~\cite{thatAnnoyingSurrogateOne}, and to accelerate anomaly detection~\cite{surrogateToSpeedUpAD}.

In contrast, we propose surrogate anomaly detection to directly learn an approximation of the regular patterns in the input data. This approach enables a more reliable measurement of deviations compared to traditional density estimation methods, particularly for complex, high-dimensional data.




\section{Theory of Surrogate Anomaly Detection}\label{sec:theory}

In engineering applications, surrogate models are frequently employed when the underlying processes are too complex to model directly. Similarly, when it comes to anomaly detection, it may be impractical to model a complex distribution directly. Here, we define a \emph{surrogate} as a model that approximately learns characteristic patterns of normal samples and identifies anomalies through deviations from these patterns.

This idea can be formalized by requiring that a learnable function $f:\mathbb{R}^d\to\mathbb{R}^k$ approximates a target pattern $g:\mathbb{R}^d\to\mathbb{R}^k$ over the set $X \subset \mathbb{R}^d$ of normal data samples:

\begin{equation}
    f({x})\approx g({x}), \;\; \forall {x}\in X
    \label{eqn:base}
\end{equation}

For example, consider a dataset where each normal sample satisfies $x_0 = x_1$. In this case, if we set $f(x) = x_0$ and $g(x) = x_1$, any significant discrepancy between $x_0$ and $x_1$ would indicate an anomaly. In practice, $f$ may be realized by training a neural network to map high-dimensional input data to a lower-dimensional latent space (with $k \ll d$), where the underlying structure of normality is more apparent.

The target pattern $g$ may be chosen in various ways. For instance, in an autoencoder $g$ is the identity function, i.e., $g_{AE}(x)=x$. However, since sufficiently expressive neural networks are universal function approximators, there are no inherent restrictions on the choice of $g$; the only requirement is that it represents a pattern that is largely invariant across normal samples.

To quantify the extent to which a sample $x$ deviates from the learned pattern, we can measure the difference between $f(x)$ and $g(x)$:

\begin{equation}
    score({x})=\|f({x})-g({x})\|
    \label{eqn:ascore}
\end{equation}

Since the goal is to ensure that normal samples conform to the learned pattern, we can minimize the aggregate deviation over the training data $X_{train}$ by employing the loss function

\begin{equation}
    \mathcal{L}=\sum_{{x}\in X_{train}} score({x})
    \label{eqn:loss}
\end{equation}

A critical observation is that while minimizing this loss drives $f(x)$ closer to $g(x)$ for normal samples, it does not directly enforce a high anomaly score for abnormal ones. Consequently, surrogate models often require additional mechanisms to avoid trivial solutions, as discussed in \cite{triplet}.

In summary, Equations~\ref{eqn:ascore} and~\ref{eqn:loss} provide a general framework for developing surrogate anomaly detection algorithms. Although any function $g$ consistent with the definition may be used, its effectiveness in yielding a well-performing anomaly detector may vary considerably.

\subsection{Surrogate Axioms}

To guide the selection of the pattern function $g$ in our surrogate model, we propose five axioms that an optimal surrogate algorithm should satisfy. We assume that a performance measure $m(f)$ (e.g., AUC-ROC) exists, which evaluates how well a model separates anomalies from normal samples.

First, note that the comparison in Equation~\ref{eqn:ascore} depends not only on the relative deviation of $f(x)$ from $g(x)$, but also on the magnitude of $\|g(x)\|$. If $\|g(x)\|$ varies significantly across samples, this may unfairly bias their anomaly score assignments.

\begin{axiom}[Scale Consistency]\label{ax:bias}
The pattern function $g$ should produce outputs of similar scale for all inputs: $\forall x_1,x_2 \in \mathbb{R}^d$ it holds that $\|g(x_1)\|\approx\|g(x_2)\|$.
\end{axiom}

An optimal surrogate must also yield similar results under identical training conditions, ensuring that any observed performance is not a mere artifact of random initialization.

\begin{axiom}[Reliable Training Procedure]\label{ax:ezy}
When learning to approximate $g$ multiple times under identical training conditions, the variance in performance should be small. For learned instances $f_1,...,f_n$ it holds that $Var(m(f_i)) \leq \delta^2$, where the constant $\delta > 0$ is as as small as possible.
\end{axiom}

It is also crucial to be robust against trivial solutions \textendash{} functions that (locally) minimize the loss $\mathcal{L}$, yet have no ability to discern between normal and anomalous samples \textendash{} since such solutions render the model useless for anomaly detection.

\begin{axiom}[Robustness to Trivial Solutions]\label{ax:minima}
There should be no trivial solution $f_{\text{trivial}}$ such that $\nabla \mathcal{L}(f_{\text{trivial}})=0$ and $f_{\text{trivial}}(x) \approx c$ for all $x \in \mathbb{R}^d$ and some constant $c \in \mathbb{R}^k$.
\end{axiom}

Hyperparameter selection is a significant challenge in anomaly detection~\cite{zhaohyper}. Thus, an optimal surrogate should exhibit stability under reasonable variations in hyperparameters, that do not fundamentally alter the model's methodological design or learning dynamics.

\begin{axiom}[Hyperparameter Invariance]\label{ax:hyper}
For any two reasonable hyperparameter sets $H_A$ and $H_B$, let $f_{H_A}$ and $f_{H_B}$ be the corresponding learned models. Then the performance difference should be bounded as $|m(f_{H_A}) - m(f_{H_B})| \leq \eta$, where $\eta > 0$ is chosen to be as small as possible.
\end{axiom}

Finally, because anomalies can be both complex and subtle, it is imperative that the surrogate model possesses sufficient expressive power. The model must be able to capture intricate patterns in the data, allowing it to accurately distinguish between normal and anomalous behavior.

\begin{axiom}[Complex Pattern Learning]\label{ax:scale}
The learnable function $f$ needs to be represented by a universal function approximator, capable of approximating any continuous function $g : \mathbb{R}^d \to \mathbb{R}^k$ to arbitrary precision on subsets of $\mathbb{R}^d$.
\end{axiom}

\subsection{Axiom Compliance}

Both of the most established deep anomaly detection paradigms that conform to our surrogate definition exhibit significant deviations from the proposed axioms,  highlighting inherent limitations in their design.

\textbf{Autoencoder:} An Autoencoder~\cite{aean} defines its surrogate usually via the identity function $g_{AE}(x)=x$, training a neural network to reconstruct its input while enforcing a compression step to prevent the trivial layer-by-layer identity mapping. However, this approach violates Axiom~\ref{ax:hyper} because the latent dimensionality must be carefully chosen, which critically affects performance. Moreover, autoencoders can converge to local minima \textendash{} such as outputting the mean of the training samples (violating Axiom~\ref{ax:minima}) \textendash{} and the lack of consistent scaling in $g$ results in biased anomaly scores (violating Axiom~\ref{ax:bias}). 


\textbf{DeepSVDD:} DeepSVDD~\cite{deepsvdd} constructs its surrogate model using a constant pattern function $g_{SVDD}(x)=c$, where $c$ is a predetermined constant, usually chosen as the mean output of the initialized network. To mitigate the risk of learning a trivial constant prediction, the method suggests avoiding bounded activation functions and removing the learnable shifts\footnote{We refer to the bias term of a neural network as "learnable shift" to reduce confusion.} from each network layer. Unfortunately, the latter restriction limits the network's capacity to learn complex patterns, thereby breaching either Axiom~\ref{ax:minima} or Axiom~\ref{ax:scale}.

\section{A Minimal Surrogate: The DEAN Model}

Given that no surrogate known to us adheres to all aforementioned axioms, we propose a novel deep learning-based approach. We observe that increased complexity in the pattern function $g$ often leads to arbitrary weighting of different samples (Axiom~\ref{ax:bias}) and intensifies challenges during training for the function $f$ that needs to be learned (Axioms~\ref{ax:ezy} and~\ref{ax:minima}). Thus, we advocate for selecting the simplest possible function $g$ that adequately identifies the essential data patterns.

Depending on the measure of complexity, one might consider $g_0(x)=0$ as the simplest option. However, this surrogate violates Axiom~\ref{ax:minima} by introducing a local minimum where every weight in the last layer becomes zero~\cite{deepsvdd}. Although such a minimum might not always be reached or could be avoided using regularization, these strategies would, in turn, compromise our remaining axioms. Instead, we propose $g_{DEAN}({x})=1$ and the surrogate generated by it. While a local minimum may still occur via the learnable shifts in the final layer, it is more manageable, as we demonstrate later (Section~\ref{sec:localminima}).

Thus, we train a neural network to output a constant value of $1$. In accordance with our framework, the loss and score functions are defined as

\begin{equation}
    \mathcal{L}=\sum_{{x}\in X_{train}} \|f({x})-1\|, \hspace{3em} score({x})=\|f({x})-q\|
    \label{eqn:losstrue}
\end{equation}
where 
\[
q=\frac{1}{\|X_{\text{train}}\|}\sum_{x_T\in X_{\text{train}}} f(x_T) \approx 1
\]

This choice of $q$ ensures that the distribution of normal samples is centered, thereby improving the robustness of the anomaly score.

Since we only learn a one-dimensional pattern with this approach, the model may, in the worst case, fix only one feature as a function of the remaining ones, which can lead to an increased number of false negatives. While one might extend $g$ to a higher-dimensional constant vector $g(x)=(1,1,\dots,1)^T$, this typically results in significantly correlated outputs across the network’s features and may violate Axiom~\ref{ax:hyper}. To address these challenges, we propose using an ensemble of surrogates. Specifically, we combine many independently trained submodels with $g_{DEAN}$ into a more effective model $F$. An integer constant, denoted by $power$, guides the aggregation of these models:
 
\begin{equation}
    score_{F}(x)=\frac{1}{\|F\|}\sum_{f_i \in F} \|f_i(x)-q_i\|^{power}
    \label{eqn:ensemble}
\end{equation}

where each $q_i$ is computed analogously to $q$. Owing to the simplicity of our surrogate, training each neural network takes only seconds, thus allowing us to combine a large number of submodels. The use of fully independent networks facilitates parallelization, reduces the correlation among learned patterns, and permits the application of ensemble methods such as feature bagging~\cite{feature-bagging} to further improve diversity and runtime consistency in high-dimensional settings. Feature bagging additionally can ensure a constant number of features for each submodel, resulting in a close-to-constant runtime for higher dimensional data.

We refer to this overall setup as \emph{DEAN} (Deep Ensemble ANomaly detection).

\subsection{DEAN Parameterization}\label{sec:hyper}

In our instantiation of DEAN, we advocate for a diverse ensemble of simple and efficient feedforward networks.

\textbf{Network Architecture:} We use a basic Multi-Layer Perceptron (MLP) with only a few hidden layers. Hidden layers are constructed with bias terms and use ReLU activations, while the output layer excludes a bias term and employs a SELU activation to mitigate the risk of dead neurons.

\textbf{Ensemble Structure:} A large ensemble size allows specialization in a variety of different patterns. For high-dimensional datasets, feature bagging is used to promote the diversity of submodels by training each on a random subset of the available features. For datasets with few features, all of them may be used to ensure critical correlations are captured.

\textbf{Power Parameter:} The $power$ parameter allows controlling the sensitivity of the aggregated anomaly score. A higher value accentuates significant deviations in one model over multiple smaller deviations across models.

\textbf{Training Configuration:} A lower-than-standard learning rate is paired with a relatively high batch size. This configuration not only stabilizes training but also encourages the network to converge towards local minima, which is beneficial for ensemble diversity.

\subsection{Axiom Compliance of DEAN}\label{sec:localminima}

DEAN is designed to fulfill all of our surrogate axioms. Since $g(x)=1$ for all $x$, Axiom~\ref{ax:bias} (Scale Consistency) is satisfied. The compliance with Axioms~\ref{ax:ezy} (Reliable Training Procedure) and \ref{ax:hyper} (Hyperparameter Invariance) is best evaluated via experimental comparisons (see Section~\ref{sec:axcexp}). We now discuss the more challenging Axioms~\ref{ax:minima} (Robustness to Trivial Solutions) and~\ref{ax:scale} (Complex Pattern Learning). This is because the trivial function $f_{trivial}({x})=0\cdot{x}+1$ perfectly minimizes the DEAN loss and is independent of its input.

\textbf{Axiom~\ref{ax:minima}:} Given its ensemble structure, DEAN inherently mitigates trivial solutions. In the event that they occur, their contribution to the ensemble anomaly score is zero since $f_{trivial}(x)=1$ for all $x$, hence $\|f_{trivial}(x)-q\|=0$.
Thus, with a sufficient number of submodels, the overall performance of DEAN remains unaffected by any individual trivial solution.

\textbf{Axiom~\ref{ax:scale}:} DeepSVDD addresses a similar issue of trivial solutions by removing learnable shifts entirely~\cite{deepsvdd}; however, this approach limits the network's capacity and violates Axiom~\ref{ax:scale}. To still mitigate this risk, while preserving expressiveness, we remove learnable shifts only from the final layer. This adjustment increases the complexity required to achieve a trivial solution, making it less likely to be reached during training, while ensuring that the network retains the ability to approximate any function (see Appendix~\ref{app:bias}).

\section{Experimental Evaluation}\label{sec:exp}

To experimentally evaluate our method, we refer to the protocol outlined in the survey paper ADBench~\cite{surveyzhao}. ADBench recommends $121$ datasets ($57$ of which are entirely uncorrelated) for benchmarking unsupervised anomaly detection algorithms, as well as a set of baseline algorithms to compare against.

\subsection{Experimental Setup}

Following the approach of ADBench, we compare DEAN against a total of $19$ state-of-the-art algorithms. This includes KNN~\cite{otherknn}, LOF~\cite{lof}, CBLOF~\cite{cblof}, Isolation Forest~\cite{ifor}, PCA~\cite{pca}, DeepSVDD~\cite{deepsvdd}, OCSVM~\cite{ocsvm}, LODA~\cite{loda}, HBOS~\cite{hbos}, COPOD~\cite{copod}, ECOD~\cite{ecod}, SOD~\cite{sod} and DAGMM~\cite{dagmm}. In addition, we consider a regular Autoencoder~\cite{aean}, as it is also a surrogate algorithm, as well as a variational autoencoder~\cite{vae} and a normalizing flow~\cite{nf} as deep learning density-based competitors. To further capture recent advances not originally considered in ADBench, we also compare against NeuTral-AD~\cite{NeuTralAD} (which leverages contrastive learning), DTE~\cite{dte} (based on diffusion models), and GOAD~\cite{goad} (which employs geometric transformations). In contrast to ADBench, all models are trained on uncontaminated data (one-class setting). 

For the parameterization of DEAN, we adhere to the guidelines proposed in Section~\ref{sec:hyper} in order to train an ensemble of $100$ submodels for $50$ epochs each, using early stopping with a patience of $10$ epochs. We adopt the following specific hyperparameter choices: A feedforward neural network with three hidden layers of $255$ neurons each. A lower-than-standard learning rate of $0.0001$ and a rather high batch size of $512$. Feature bagging with 200 random features per model for datasets containing at least 200 features. A power parameter set to $9$, emphasizing pronounced deviations in anomaly detection. We consider variations in ensemble size and other hyperparameters in Sections~\ref{sec:runtime} and~\ref{sec:axcexp}. Detailed information regarding the implementation and parameterization of the compared methods can be found in our code repository. 

\subsection{Anomaly Detection Performance}

\begin{table*}[ht]
\centering
\resizebox{\textwidth}{!}{
\begin{tabular}{lcc}
\hline
\textbf{Algorithm} & \textbf{AUC-ROC (all)} & \textbf{AUC-ROC (larger half of datasets)} \\
\hline
\multirow{22}{*}{
\begin{tabular}{@{}l@{}}
\vspace{1.5em} \\
\blueSquare DEAN \vspace{0.08em}\\
\blueSquare Autoencoder~\cite{aean} \vspace{0.08em}\\
\blueSquare DeepSVDD~\cite{deepsvdd} \vspace{0.08em}\\
\greenSquare PCA~\cite{pca} \vspace{0.08em}\\
\blueTriangleReverse NF~\cite{nf} \vspace{0.08em}\\
\blueTriangleReverse DTE~\cite{dte} \vspace{0.08em}\\
\blueTriangleReverse GOAD~\cite{goad} \vspace{0.08em}\\
\blueTriangleReverse DAGMM~\cite{dagmm} \vspace{0.08em}\\
\blueTriangleReverse NeuTral~\cite{NeuTralAD} \vspace{0.08em}\\
\blueTriangleReverse VAE~\cite{vae} \vspace{0.08em}\\
\greenTriangleReverse SOD~\cite{sod} \vspace{0.08em}\\
\greenTriangleReverse OCSVM~\cite{ocsvm} \vspace{0.08em}\\
\greenTriangleReverse LOF~\cite{lof} \vspace{0.08em}\\
\greenTriangleReverse LODA~\cite{loda} \vspace{0.08em}\\
\greenTriangleReverse KNN~\cite{otherknn} \vspace{0.08em}\\
\greenTriangleReverse IForest~\cite{ifor} \vspace{0.08em}\\
\greenTriangleReverse HBOS~\cite{hbos} \vspace{0.08em}\\
\greenTriangleReverse ECOD~\cite{ecod} \vspace{0.08em}\\
\greenTriangleReverse COPOD~\cite{copod} \vspace{0.08em}\\
\greenTriangleReverse CBLOF~\cite{cblof} 
\end{tabular}
} &
\adjustbox{valign=t}{\includegraphics[width=0.6\textwidth]{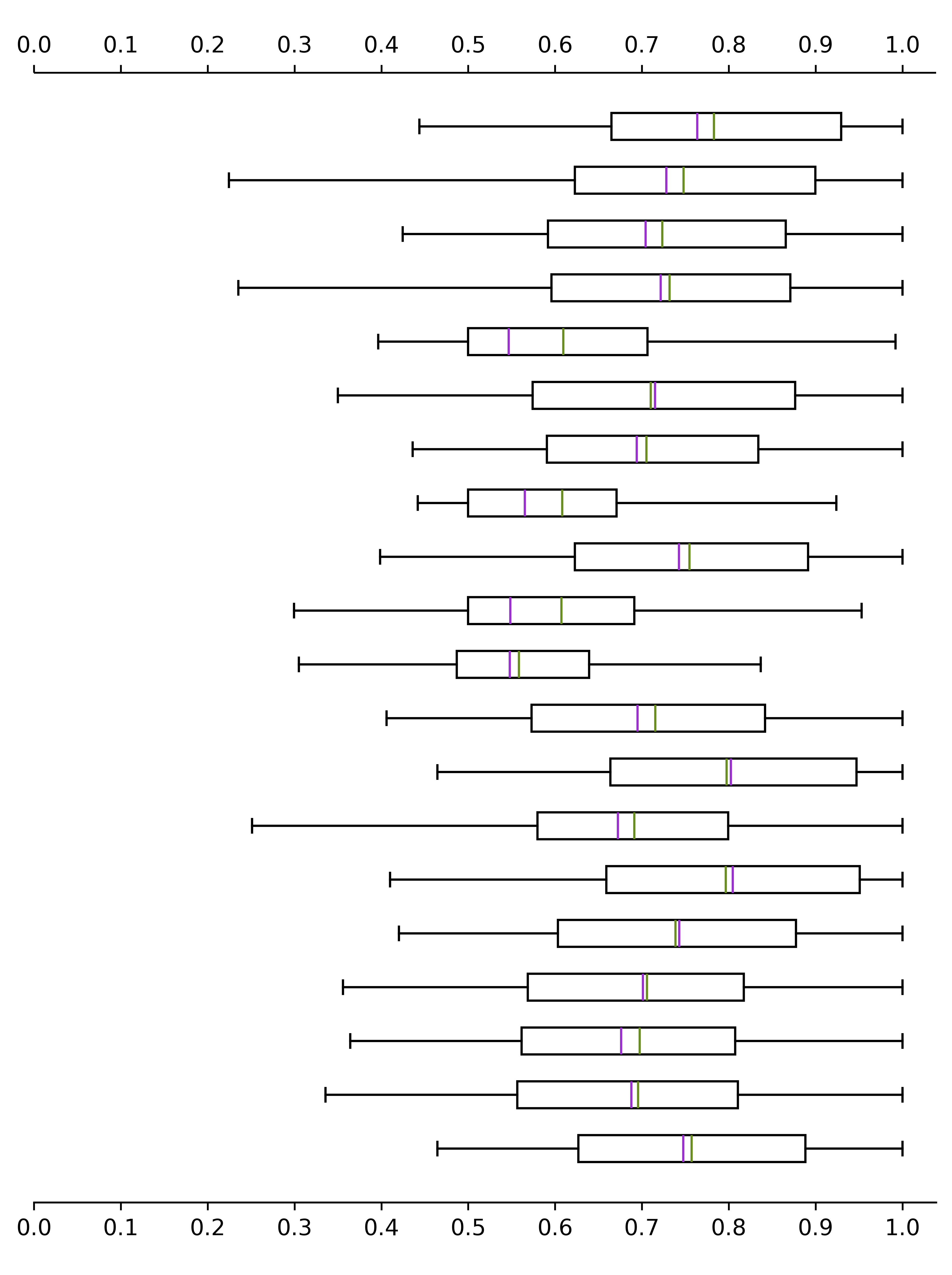}} &
\adjustbox{valign=t}{\includegraphics[width=0.6\textwidth]{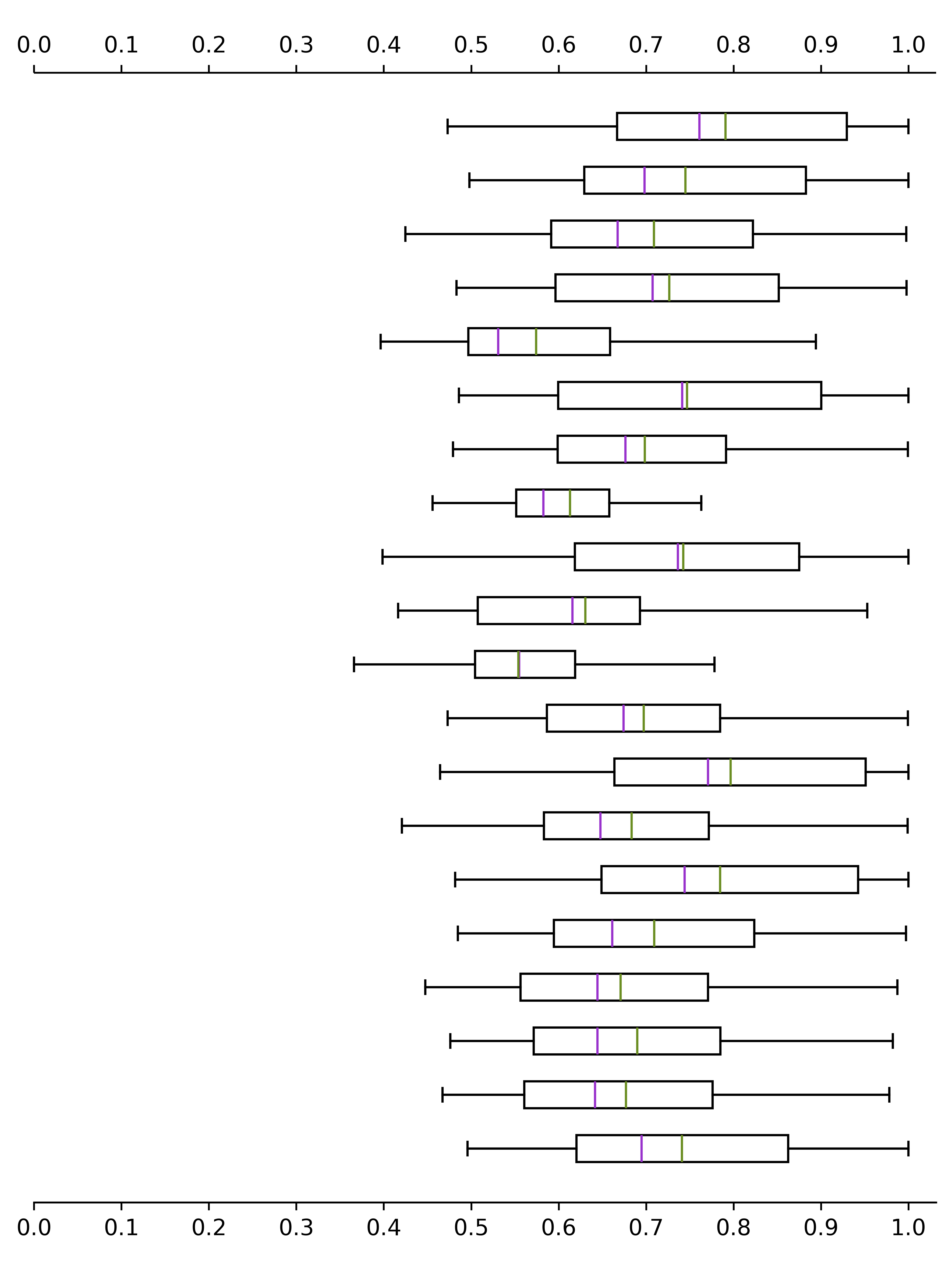}} \\
\hline
\end{tabular}
}
\caption{Distribution of AUC-ROC performance for all evaluated algorithms. Deep learning models (blue) and shallow models (yellow) are differentiated by surrogate status (squares for surrogates, triangles for non-surrogates). Mean and median values are shown in green and purple, respectively.}
\label{table:performance_metrics}
\end{table*}

Our primary performance evaluation metric is the Area Under the Receiver Operating Characteristic (AUC-ROC). For additional insight, we provide a complementary evaluation based on the Area Under the Precision-Recall Curve (AUC-PR), alongside individual results for each dataset, in Appendix~\ref{app:pr} and~\ref{app:raw}.

A summarization of the observed AUC-ROC performance is given in Table~\ref{table:performance_metrics}. Consistent with the results from ADBench, our analysis shows that no method outperforms all others in a statistically significant manner across all datasets. Our algorithm performs highly competitively when averaged across all datasets and is only slightly outperformed by KNN and LOF. These competitors do not scale well to the more challenging datasets. Thus, when only considering the larger half of the datasets studied here, the median performance of DEAN is higher than all competitors considered. 

\begin{figure}[h]
  \centering
  \begin{subfigure}[b]{0.48\linewidth}
    \centering
    \includegraphics[width=\linewidth]{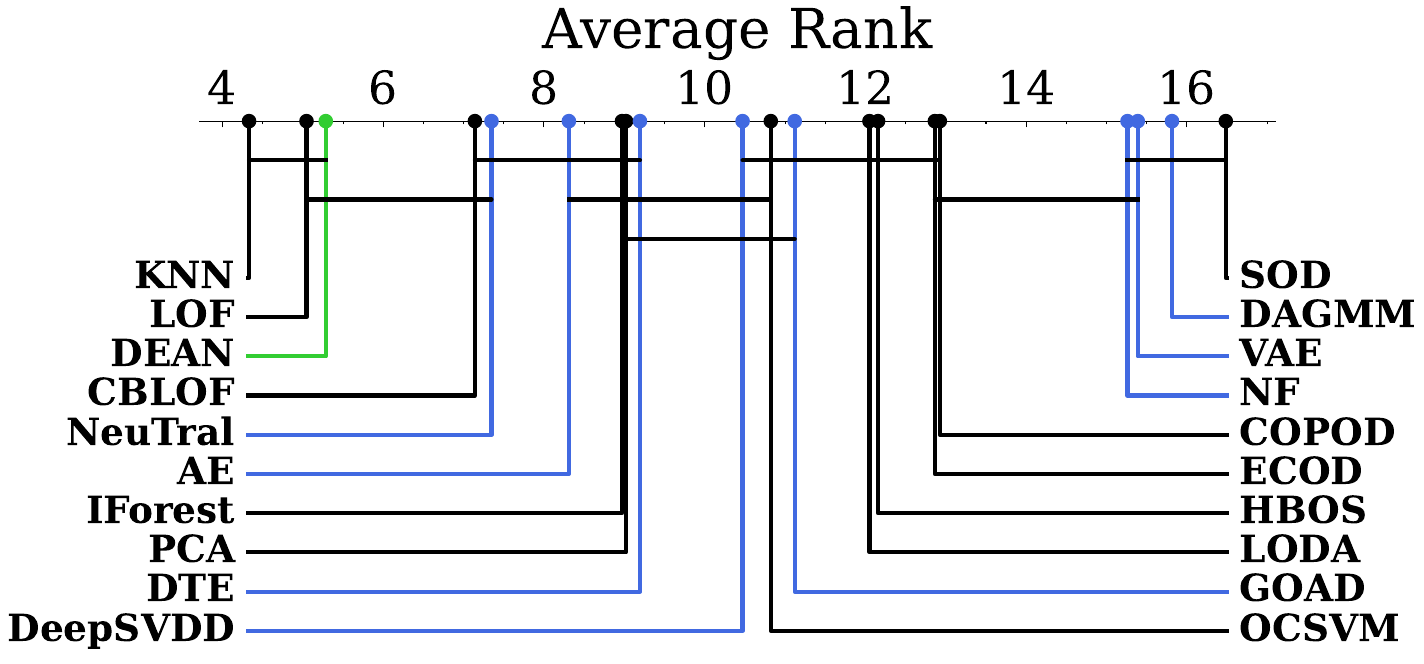}
    \caption{Using all ADBench datasets.}
    \label{fig:cd}
  \end{subfigure}
  \hfill
  \begin{subfigure}[b]{0.48\linewidth}
    \centering
    \includegraphics[width=\linewidth]{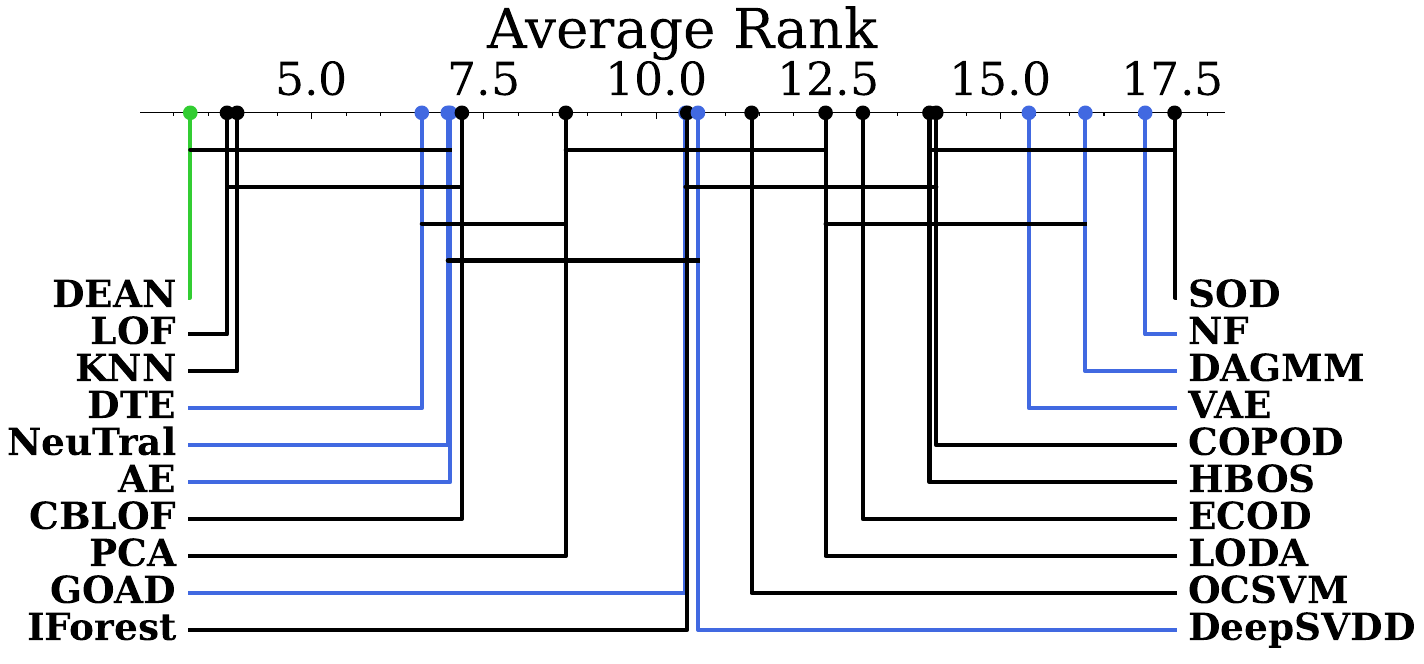}
    \caption{Using only the larger half of datasets.}
    \label{fig:cd2}
  \end{subfigure}
  \caption{Critical difference diagrams comparing the AUC-ROC performance. A lower rank indicates better performance, while algorithms with no statistically significant differences are connected by a horizontal line. DEAN is depicted in green, other deep learning algorithms in blue.}
  \label{fig:cd_combined}
\end{figure}

To further illustrate our findings and provide a concise ranking of performance, we provide the critical difference diagrams in Figure~\ref{fig:cd_combined}. In these diagrams, a Friedman test~\cite{friedman} is used to determine if significant differences exist between algorithm performances (measured in AUC-ROC), and algorithms with no significant differences are connected using a Wilcoxon test~\cite{wilcoxon}. We consider p-values below $p\le5\%$ after Bonferroni-Holm correction~\cite{correction} to be significant.

Notably, DEAN performs significantly stronger than every other surrogate or deep learning algorithm, with the exception of NeuTral (see Figure~\ref{fig:cd}). Similar to ADBench, widely recognized shallow algorithms such as KNN, LOF, and CBLOF remain strong competitors. Our algorithm outperforms CBLOF, but does not quite achieve the same average rank as KNN and LOF. We attribute this outcome to the fact that benchmark datasets are often low-dimensional, contain a large number of samples, and exhibit anomalies that are relatively simple in nature. This combination favors distance-based methods, as differences in local densities are more pronounced; however, they may not capture the complexity encountered in real-world applications. Moreover, the lazy learning paradigm intrinsic to KNN and LOF, which necessitates the retention of training instances during inference, is well known to scale badly to large, high-dimensional datasets~\cite{charu-book-outlier-analysis}.

For instance, when considering only the larger half of the datasets, DEAN emerges as the best fit, outperforming every competitor (see Figure~\ref{fig:cd2}). This performance advantage, coupled with its scalability and robust learning framework, makes DEAN particularly well suited for advanced tasks where the expressive power of neural networks is needed without introducing unnecessary complexity.

\subsection{Runtime and Ensemble Analysis}\label{sec:runtime}

\begin{figure}[h]
  \centering
  \begin{subfigure}[b]{0.54\linewidth}
    \centering
    \includegraphics[width=\linewidth]{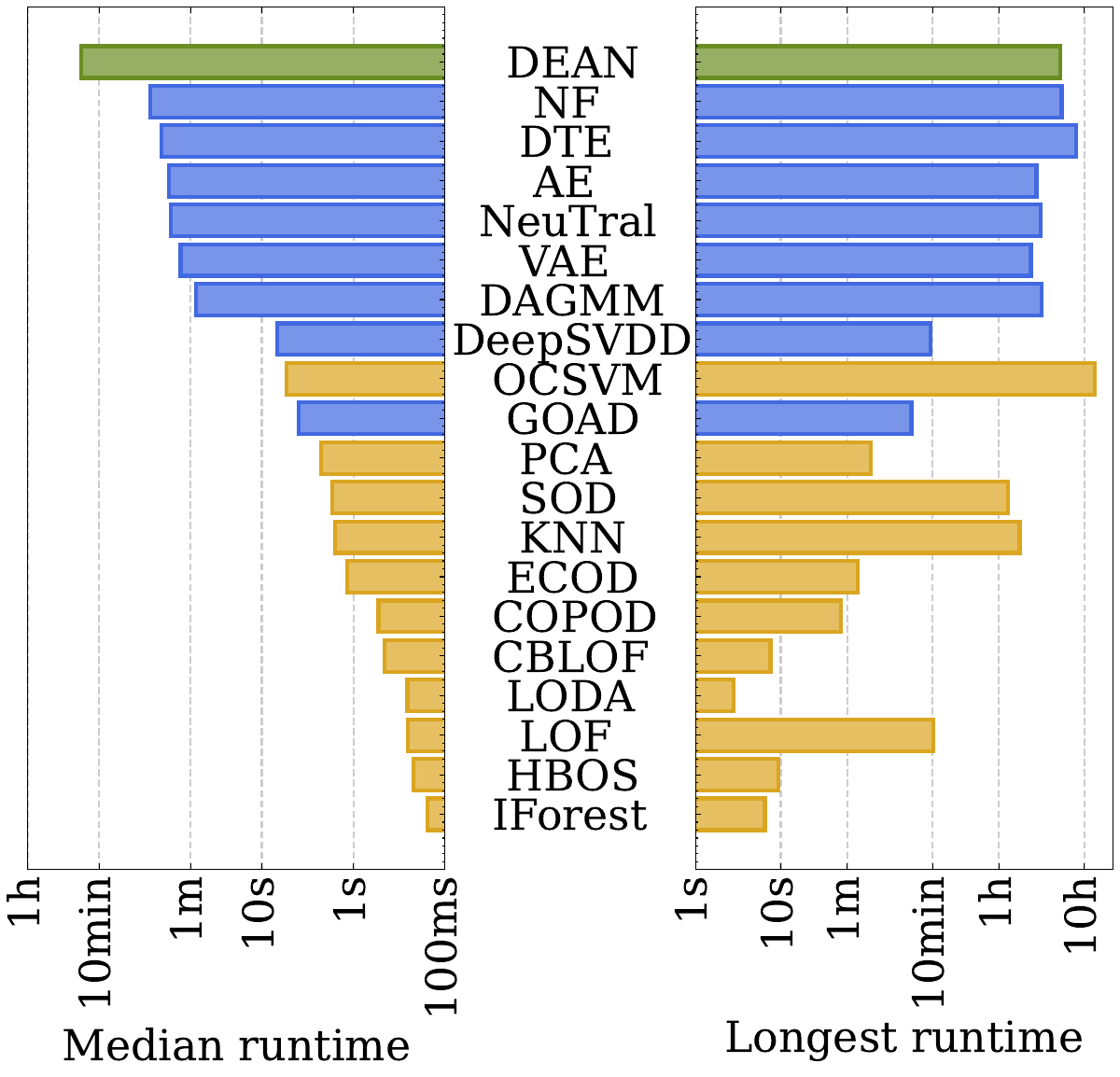}
    \caption{Total Runtime comparison.}
    \label{fig:time}
  \end{subfigure}
  \hfill
  \begin{minipage}[b]{0.42\linewidth}
    \begin{subfigure}[b]{\linewidth}
      \centering
      \includegraphics[width=\linewidth]{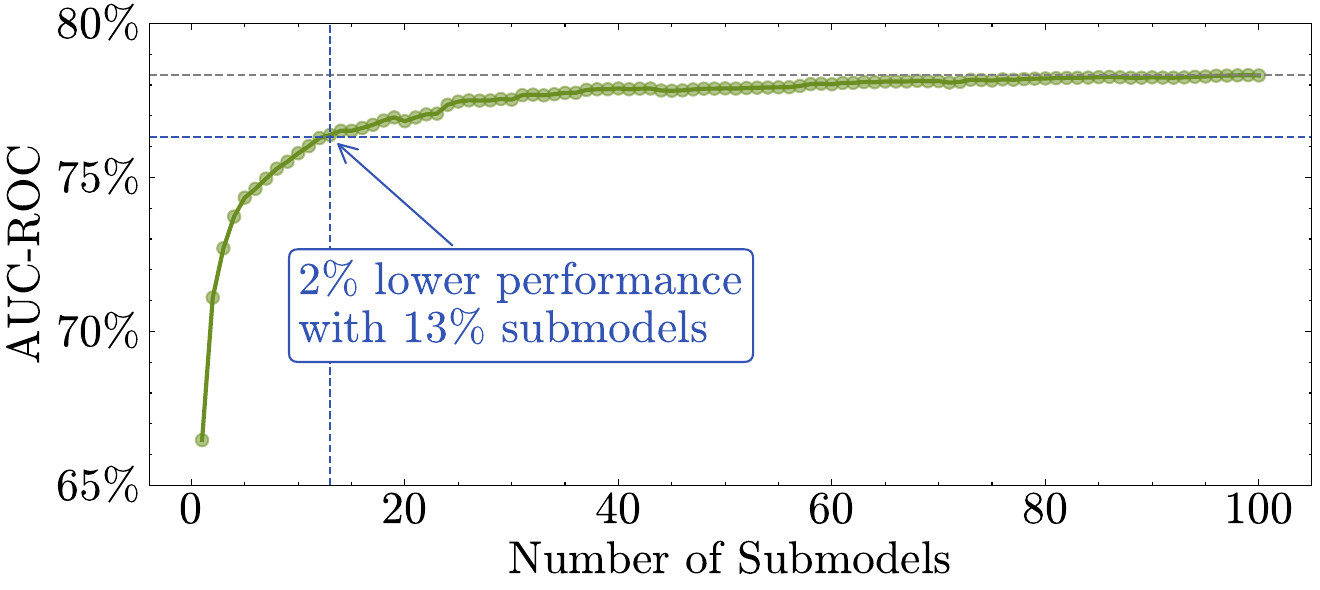}
      \caption{Ensemble performance growth.}
      \label{fig:growth}
    \end{subfigure}
    \vspace{0.em} 
    
    \begin{subfigure}[b]{\linewidth}
      \centering
      \includegraphics[width=\linewidth]{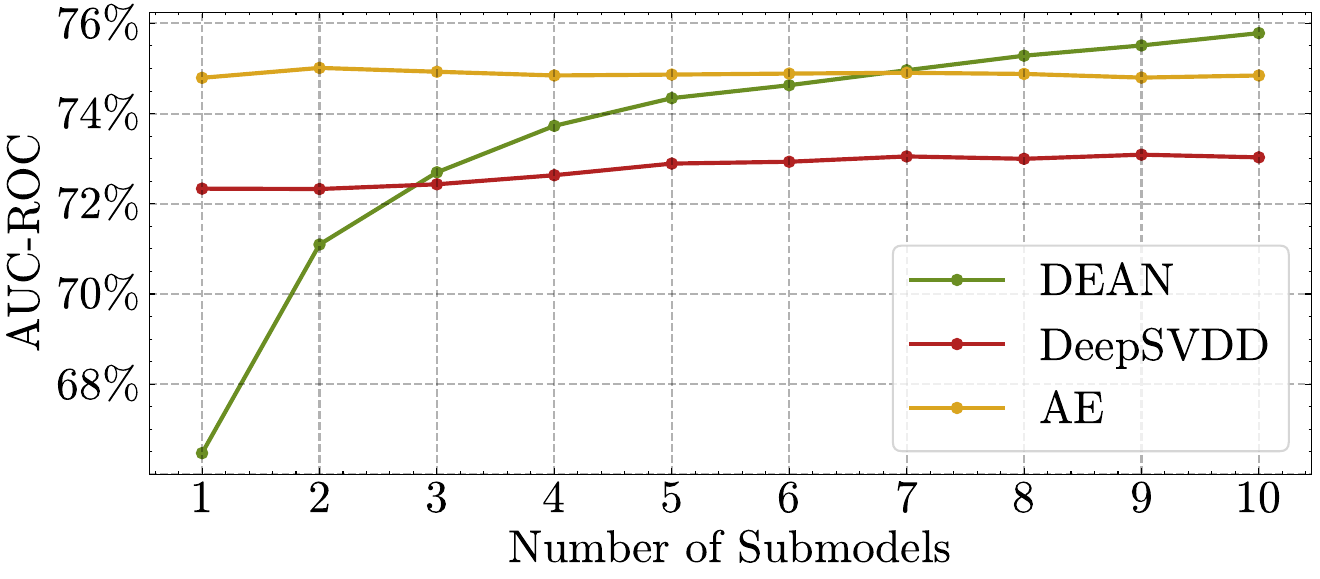}
      \caption{Competitor ensembles.}
      \label{fig:ens}
      \vspace{2.5em}
    \end{subfigure}
  \end{minipage}
  \caption{(a) Overall runtime overview across all datasets; DEAN is depicted in green, other deep learning algorithms in blue, and shallow algorithms in yellow. (b) and (c) average AUC-ROC performance changes with varying ensemble size.}
  \label{fig:combined}
\end{figure}

Figure~\ref{fig:combined} provides an overview of our runtime measurements and the impact of using (larger) ensembles on the performance of deep learning-based surrogate models. To this end, Figure~\ref{fig:time} reports both the median and maximum runtimes across datasets to account for the substantial variability in dataset sizes. All experiments were conducted on a system running Ubuntu 22.04.3 LTS, powered by an Intel\textsuperscript{\textregistered} Xeon\textsuperscript{\textregistered} w9-3495X processor with a base clock of approximately 3400~MHz and turbo boost frequencies up to around 4500~MHz and with 495~GB of RAM available. The runtime measurements were obtained using single-core execution for each algorithm to ensure a fair comparison.

As expected, deep learning methods generally exhibit longer runtimes, with the DEAN ensemble showing a median runtime of approximately $15$ minutes per dataset. Notably, for a DEAN ensemble comprising $100$ submodels, this corresponds to an average training time of less than $9$ seconds per submodel. However, it is important to emphasize that deep learning methods are particularly well-suited for GPU acceleration, and DEAN, as an ensemble method of independently optimized submodels, can be almost perfectly parallelized. Furthermore, due to the use of feature bagging, the worst-case runtime scenario is significantly mitigated, with most deep learning approaches requiring comparable or even longer runtimes.

Naturally, the runtime is also rather sensitive to the number of submodels. As illustrated in Figure~\ref{fig:growth}, while increasing the number of submodels improves performance, the relationship is non-linear. Using only 13 submodels results in an average performance that is merely $2\%$ lower than that achieved with 100 submodels, yet it requires approximately $87\%$ less training time.
At the same time, the continued performance improvement with additional submodels reflects the high variance of the individual models used in DEAN, incentivized by the simplicity of the submodels. In contrast, Figure~\ref{fig:ens} shows that ensembles based on Autoencoder or DeepSVDD methods exhibit nearly constant performance, likely due to their complex, less diverse submodel characteristics.

\subsection{Evaluation of Axiom Compliance}\label{sec:axcexp}
\begin{figure}[h]
  \centering
  \begin{subfigure}[b]{0.46\linewidth}
    \centering
    \setlength{\belowcaptionskip}{-25pt}\includegraphics[width=\linewidth]{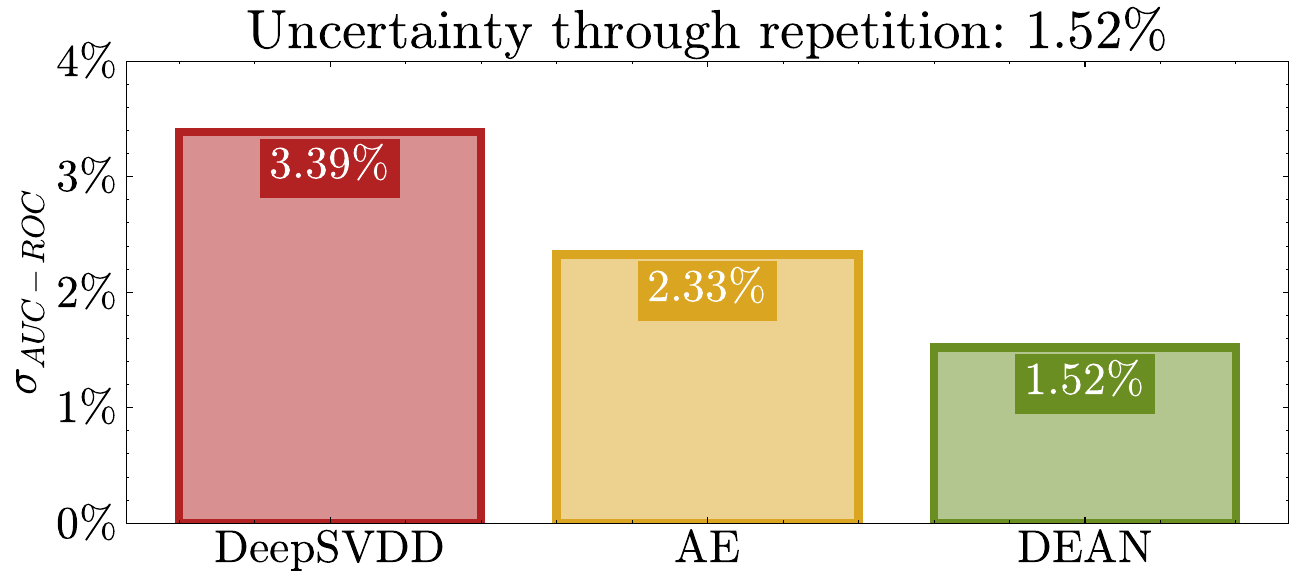}
    \label{fig:rep}
    \vspace{0.5em}
  \end{subfigure}
  \hfill
  \begin{subfigure}[b]{0.48\linewidth}
    \centering
    \setlength{\belowcaptionskip}{-25pt}\includegraphics[width=\linewidth]{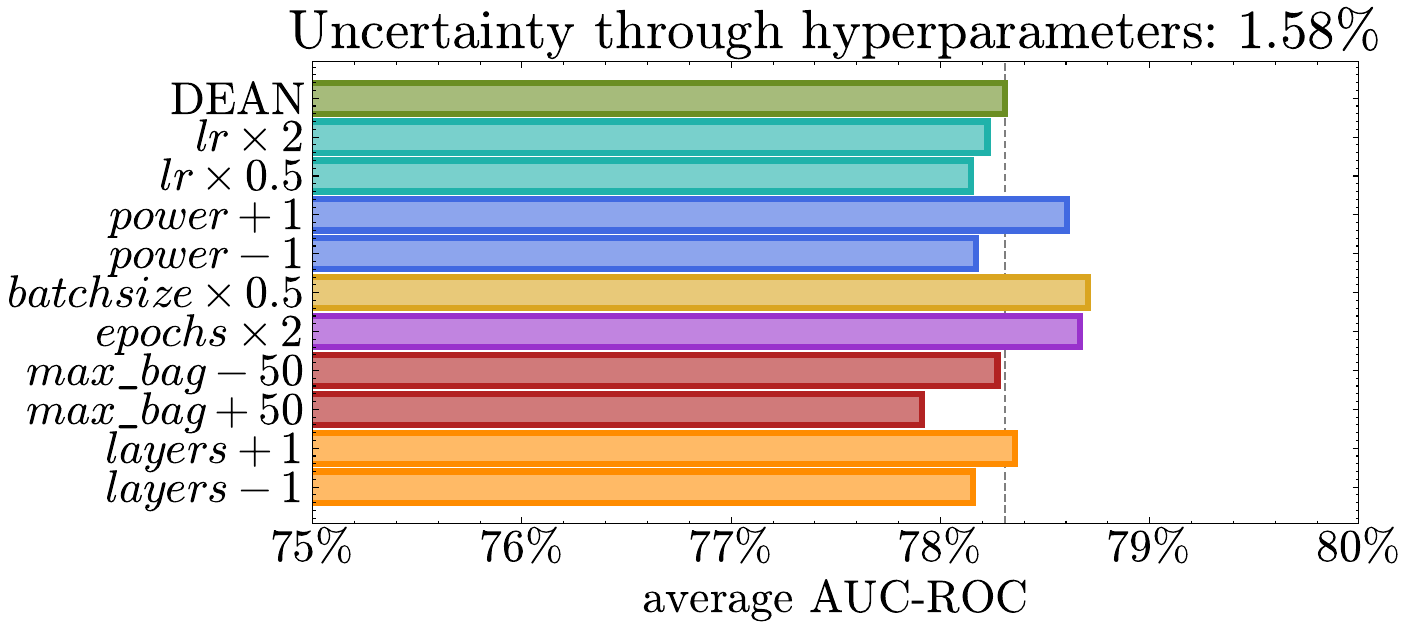}
    \label{fig:hyp}
  \end{subfigure}
  \caption{Left: Repetition uncertainty for various surrogate algorithms, Right: DEAN performance with varied hyperparameters and the resulting uncertainty.}
  \label{fig:ax_combined}
\end{figure}

Compliance with Axioms~\ref{ax:ezy} and~\ref{ax:hyper} is difficult to assess theoretically, therefore we evaluate these properties experimentally on the same datasets. For Axiom~\ref{ax:ezy}, Figure~\ref{fig:ax_combined} demonstrates that the repetition uncertainty of DEAN \textendash{} calculated as the standard deviation across 10 runs per datasets and then averaged \textendash{} is lower than that observed for other surrogate deep learning algorithms under identical training conditions. For Axiom~\ref{ax:hyper}, we present DEAN's performance when evaluated with modified hyperparameter sets.
Since the average performances are nearly equal, one may argue that the influence of such modifications is negligible. This is also in stark contrast to DeepSVDD~\cite{surveyzhao} and Autoencoder~\cite{AEstudies} behavior.

\section{Anomaly Detection Beyond Benchmarks}\label{sec:mods}

While DEAN reliably achieves competitive performance with an easy-to-configure parameterization, real-world applications often demand flexibility beyond standard benchmarks. The simplicity of our submodels and the inherent ensemble structure render DEAN highly adaptable.

For instance, the ensemble structure facilitates explainability via Shapley values~\cite{deanshap}; feature bagging helps mitigate the high computational cost of such methods. In addition, the ensemble character natively supports a distributed implementation through federated learning~\cite{federatedlearning}. The ensemble also enables the incorporation of secondary requirements, such as robustness against adversarial attacks by pruning or reweighting less robust submodels~\cite{dean}.
The simplicity of each submodel also permits modifications in the training procedure to incorporate additional information~\cite{superdean}, such as in semi-supervised anomaly detection~\cite{ruffsemisuper} or outlier exposure~\cite{outlierexposure}. Moreover, employing different machine learning models within the DEAN framework could yield lightweight variants suitable for resource-constrained devices such as IoT systems~\cite{sean}.

To summarize, we see three major ways DEAN can be adapted: (1) altering the selection of submodels, (2) adjusting the ensemble weighting, and (3) modifying the submodel training procedure. As a proof-of-concept, we apply all three adaptations for the task of fair anomaly detection~\cite{deep_fair_svdd} (see Appendix~\ref{app:bias}).

\section{Conclusion}

In this paper, we present the first systematic study of surrogate models for anomaly detection and establish a comprehensive framework for constructing such models from mathematical functions. We propose five axioms that any optimal surrogate anomaly detection algorithm should satisfy and employ these axioms to develop DEAN, a novel algorithm that meets all of them.

An extensive evaluation demonstrates that it not only performs competitively \textendash{} particularly excelling over other deep learning-based methods and alternative surrogates \textendash{} but also offers exceptional adaptability. In the future, we believe that the axiomatic design of DEAN, based on an ensemble of simple submodels, can furthermore facilitate straightforward modifications to enhance secondary anomaly detection goals, like explainability, adversarial robustness, or fairness.


%
%
%
\bibliographystyle{splncs04}
\bibliography{refs,icml,new,auto}

\appendix
\section{Importance of Learnable Shifts}\label{app:bias}

Trivial solutions are a common problem also for DeepSVDD~\cite{deepsvdd}. Namely when the last layer learns a zero multiplicative weight, but the learnable shift is equal to the desired $\vec{c}$. To combat this, Ruff et. al. propose to remove the learnable shifts entirely. And while this certainly helps in making this shift impossible, it also limits how complicated a function can be learned by the neural network~\cite{universalapproxtheorem}.

\begin{figure}[h]
  \centering
  \includegraphics[width=\linewidth]{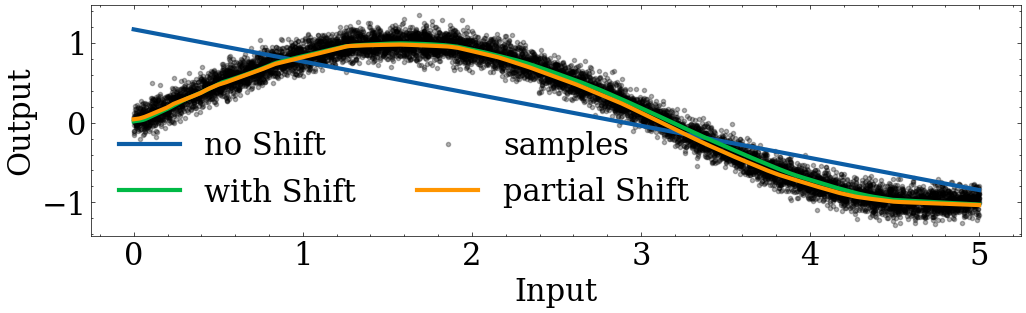}
  \caption{Given complicated alinear data, the functions learned by three neural networks with relu activations are shown. The network without learnable shifts cannot capture the structure of the underlying data, while both a network with learnable shifts in each layer and a network with learnable shifts in all layers except the last can describe the alinearity.}
  \label{fig:sinfit}
\end{figure}

We show this in Figure \ref{fig:sinfit}, where we task neural networks to approximate a simple sinus curve. Here, we use neural networks with three layers of $100$ nodes and relu activation in each hidden layer. The three networks differ only by the learnable shifts they use. While the network with learnable shifts (green) is clearly able to approximate the sinus curve, the version without learnable shifts (blue) is not able to do so. And since real anomaly representations can be much more complicated than such a simple sinus curve, we do not think that limiting the neural network complexity is a reasonable choice.

Instead, we use other methods to remove the trivial solution of a constant network. This also includes using learnable shifts in each hidden layer but not in the output layer. This setup is still able to approximate complicated functions, as is shown in orange in Figure \ref{fig:sinfit}.

\section{DEAN-Fair}\label{app:fair}

To illustrate the adaptability of DEAN (see Section 6), we demonstrate its modification for improved fairness on a toy example using the COMPAS dataset~\cite{compas}. In this context, we consider recidivism risk as the anomaly and employ fairness as a critical performance metric. The COMPAS dataset, which contains risk scores along with demographic and criminal history features, is widely used for evaluating such algorithmic fairness. 

\subsection{Setup}

For our fairness evaluation, we compute the AUC-ROC separately for two subgroups defined by a protected attribute (age, binarized with a threshold at $25$ years) and measure the deviation between them to showcase how DEAN can be guided towards equal treatment across different demographic groups in general. We chose the AUC-ROC since it is a metric invariant to the fraction of anomalous samples and also handles non-binary anomaly scores. An ideal fairness score is $0.5$, indicating no performance difference between groups.
For this, we propose three adaptation strategies to improve fairness.



\textbf{1. Modified Loss Function:}  
We add a fairness regularization term to the original loss:
\begin{equation}
    L = \sum_{\vec{x}\in X_{\text{train}}} \|f(\vec{x}) - 1\| + \theta \cdot L_{\text{fair}}
    \label{eqn:losswithfair}
\end{equation}
where
\begin{equation}
    L_{\text{fair}} = \frac{\|L_1 - L_0\|}{\|L_1\| + \|L_0\|}
    \label{eqn:lossfair}
\end{equation}
and
\begin{equation}
    L_{1/0} = \frac{1}{\|X_{\text{(un)protected}}\|}\sum_{\vec{x}\in X_{\text{(un)protected}}} f(\vec{x})
\end{equation}

Here, $L_1$ and $L_0$ denote the mean outputs for the unprotected and protected groups, respectively, and we set $\theta = 0.1$.

\textbf{2. Submodel Pruning:}  
In this approach, we iteratively remove the submodel that exhibits the greatest unfairness in a greedy manner. We test pruning rates of $1\%$, $5\%$, and $10\%$ of the ensemble.

\textbf{3. Non-uniform Weighting:}
We assign different weights to submodels in the ensemble to maximize fairness. Due to the non-continuous nature of this optimization, we employ an evolutionary algorithm to determine the optimal weights.

\subsection{Results}

Table~\ref{tab:fairness} summarizes the AUC-ROC performance and fairness (measured as the deviation from $0.5$) for each method. Each experiment is repeated five times to obtain uncertainty estimates.


\begin{table}[]
    \centering
    \setlength{\tabcolsep}{12pt} 
\begin{tabular}{lrr}
\hline
 Adjustment              & AUC-ROC           & Fairness          \\
\hline
 Baseline                & $0.583 \pm 0.003$ & $0.644 \pm 0.020$ \\
 Loss function           & $0.594 \pm 0.012$ & $0.453 \pm 0.080$ \\
 Pruning ($1\%$)         & $0.583 \pm 0.003$ & $0.625 \pm 0.019$ \\
 Pruning ($5\%$)         & $0.577 \pm 0.003$ & $0.555 \pm 0.015$ \\
 Pruning ($10\%$)        & $0.574 \pm 0.003$ & $0.506 \pm 0.014$ \\
 Non-uniform weighting   & $0.566 \pm 0.004$ & $0.520 \pm 0.011$ \\
\hline
\end{tabular}
    \caption{AUC-ROC performance and fairness deviation on the COMPAS dataset for various fairness adaptations of DEAN. Notice that the performance is better the higher the value is, while the fairness is optimal at $0.5$.}
    \label{tab:fairness}
\end{table}

The baseline model exhibits a fairness deviation of over $14\%$, indicating a significant bias. With as little as $1\%$ pruning, fairness improves, and pruning $10\%$ of the submodels nearly eliminates the bias (deviation of only $0.6\%$, within experimental uncertainty), albeit with a slight reduction in overall performance (approximately $1\%$ drop). Non-uniform weighting yields a more pronounced performance drop ($1.7\%$) and a moderate fairness improvement ($2\%$ deviation). Notably, the modified loss function further increases performance by about $1.1\%$ but overshoots fairness slightly, resulting in a $4.7\%$ deviation.

Overall, these experiments confirm that the DEAN framework can be effectively adapted to enhance fairness, demonstrating its versatility and potential for broader real-world applications.



\clearpage
\section{Performance Result Plots with AUC-PR}\label{app:pr}
Since our results are very similar whether we use AUC-ROC or AUC-PR, we only state most of our results in AUC-ROC and add the alternative plots here.

\begin{figure}[h!]
  \centering
  \includegraphics[width=\linewidth]{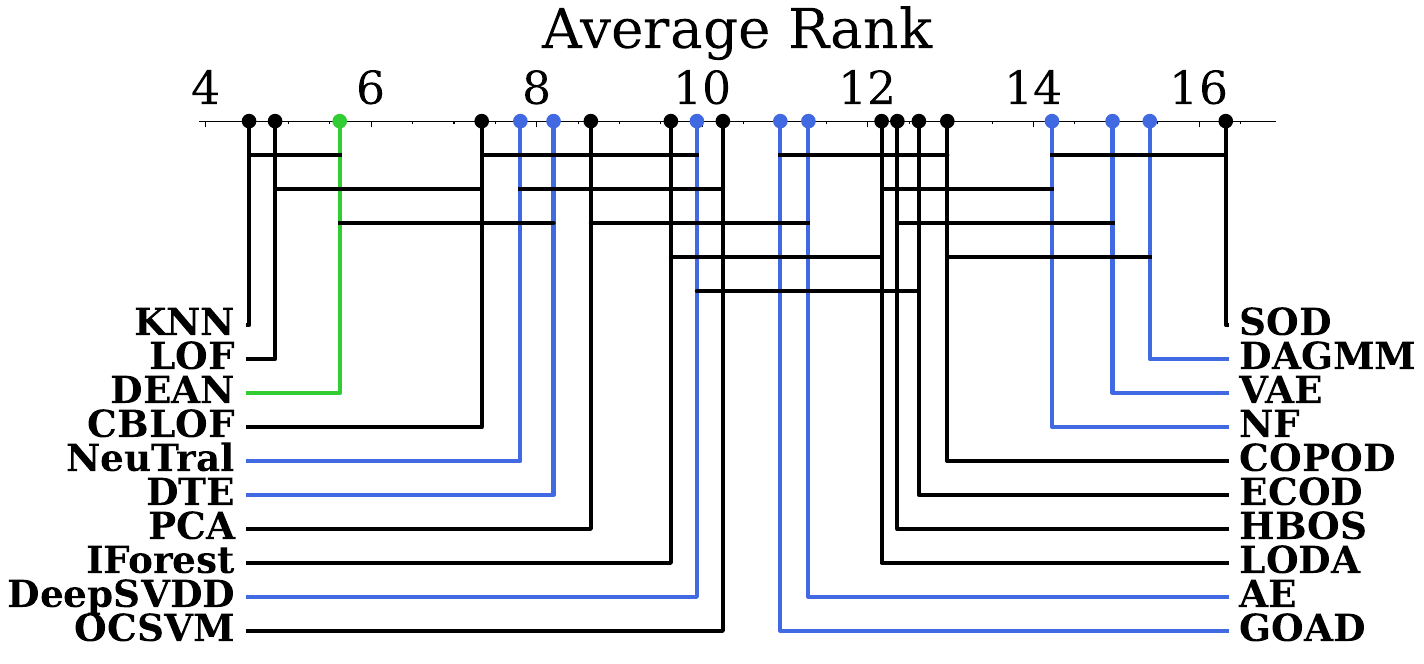}
  \caption{Critical difference diagrams comparing the AUC-PR performance. A
lower rank indicates better performance, while algorithms with no statistically
significant differences are connected by a horizontal line. DEAN is depicted in
green, other deep learning algorithms in blue. Pendant to Figure~2a in Section~5.2.}
  \label{fig:ycd}
\end{figure}

\begin{figure}[h!]
  \centering
  \includegraphics[width=\linewidth]{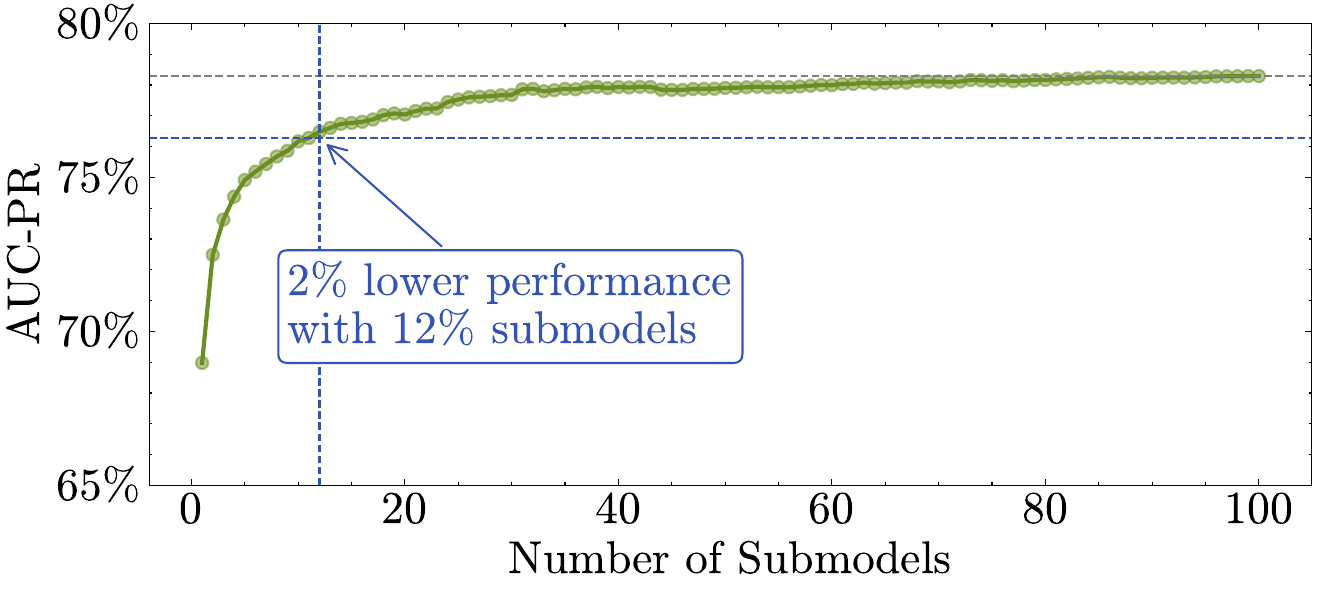}
  \caption{AUC-PR performance changes with varying ensemble size, for DEAN. It reaches $2\%$ less performance with the first $12\%$ (instead of $13\%$ for AUC-ROC) of submodels. Pendant to Figure~3b in Section~5.3.}
  \label{fig:ather_growth}
\end{figure}

Table~\ref{table:performance_metrics_pr} gives an overview of the performance for all evaluated algorithms across all datasets when using AUC-PR instead of AUC-ROC.
Figure \ref{fig:ycd} shows the critical difference plot when we use AUC-PR instead of AUC-ROC to compare the performance of algorithms. Additionally, Figure \ref{fig:ather_growth} shows the AUC-PR score as a function of the submodels used.

\begin{table*}[h!]
\centering
\resizebox{\textwidth}{!}{
\begin{tabular}{lcc}
\hline
\textbf{Algorithm} & \textbf{AUC-PR (all)} & \textbf{AUC-PR (larger half of datasets)} \\
\hline
\multirow{22}{*}{
\begin{tabular}{@{}l@{}}
\vspace{1.5em} \\
\blueSquare DEAN \vspace{0.08em}\\
\blueSquare Autoencoder~\cite{aean} \vspace{0.08em}\\
\blueSquare DeepSVDD~\cite{deepsvdd} \vspace{0.08em}\\
\greenSquare PCA~\cite{pca} \vspace{0.08em}\\
\blueTriangleReverse NF~\cite{nf} \vspace{0.08em}\\
\blueTriangleReverse DTE~\cite{dte} \vspace{0.08em}\\
\blueTriangleReverse GOAD~\cite{goad} \vspace{0.08em}\\
\blueTriangleReverse DAGMM~\cite{dagmm} \vspace{0.08em}\\
\blueTriangleReverse NeuTral~\cite{NeuTralAD} \vspace{0.08em}\\
\blueTriangleReverse VAE~\cite{vae} \vspace{0.08em}\\
\greenTriangleReverse SOD~\cite{sod} \vspace{0.08em}\\
\greenTriangleReverse OCSVM~\cite{ocsvm} \vspace{0.08em}\\
\greenTriangleReverse LOF~\cite{lof} \vspace{0.08em}\\
\greenTriangleReverse LODA~\cite{loda} \vspace{0.08em}\\
\greenTriangleReverse KNN~\cite{otherknn} \vspace{0.08em}\\
\greenTriangleReverse IForest~\cite{ifor} \vspace{0.08em}\\
\greenTriangleReverse HBOS~\cite{hbos} \vspace{0.08em}\\
\greenTriangleReverse ECOD~\cite{ecod} \vspace{0.08em}\\
\greenTriangleReverse COPOD~\cite{copod} \vspace{0.08em}\\
\greenTriangleReverse CBLOF~\cite{cblof} 
\end{tabular}
} &
\adjustbox{valign=t}{\includegraphics[width=0.6\textwidth]{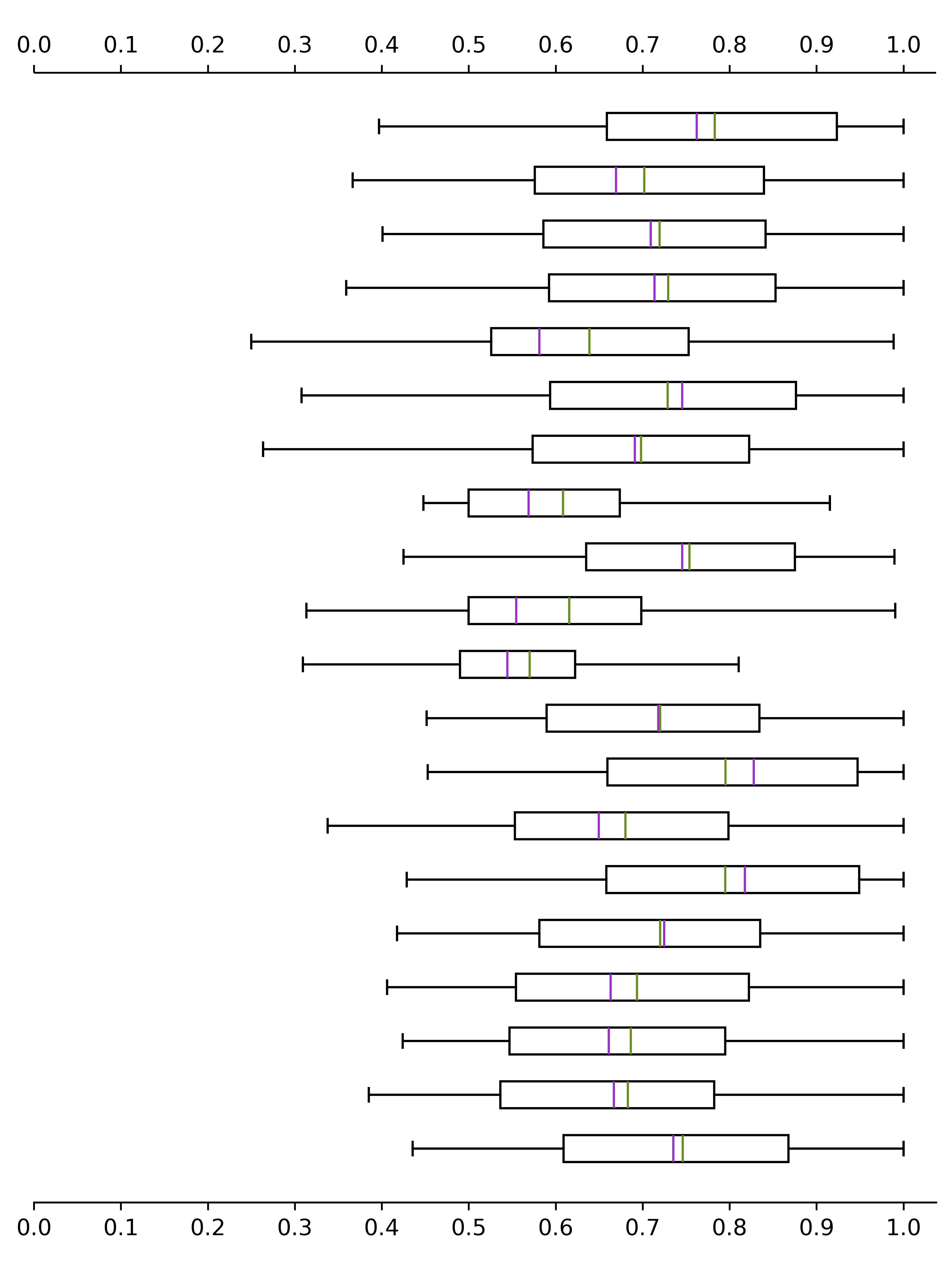}} &
\adjustbox{valign=t}{\includegraphics[width=0.6\textwidth]{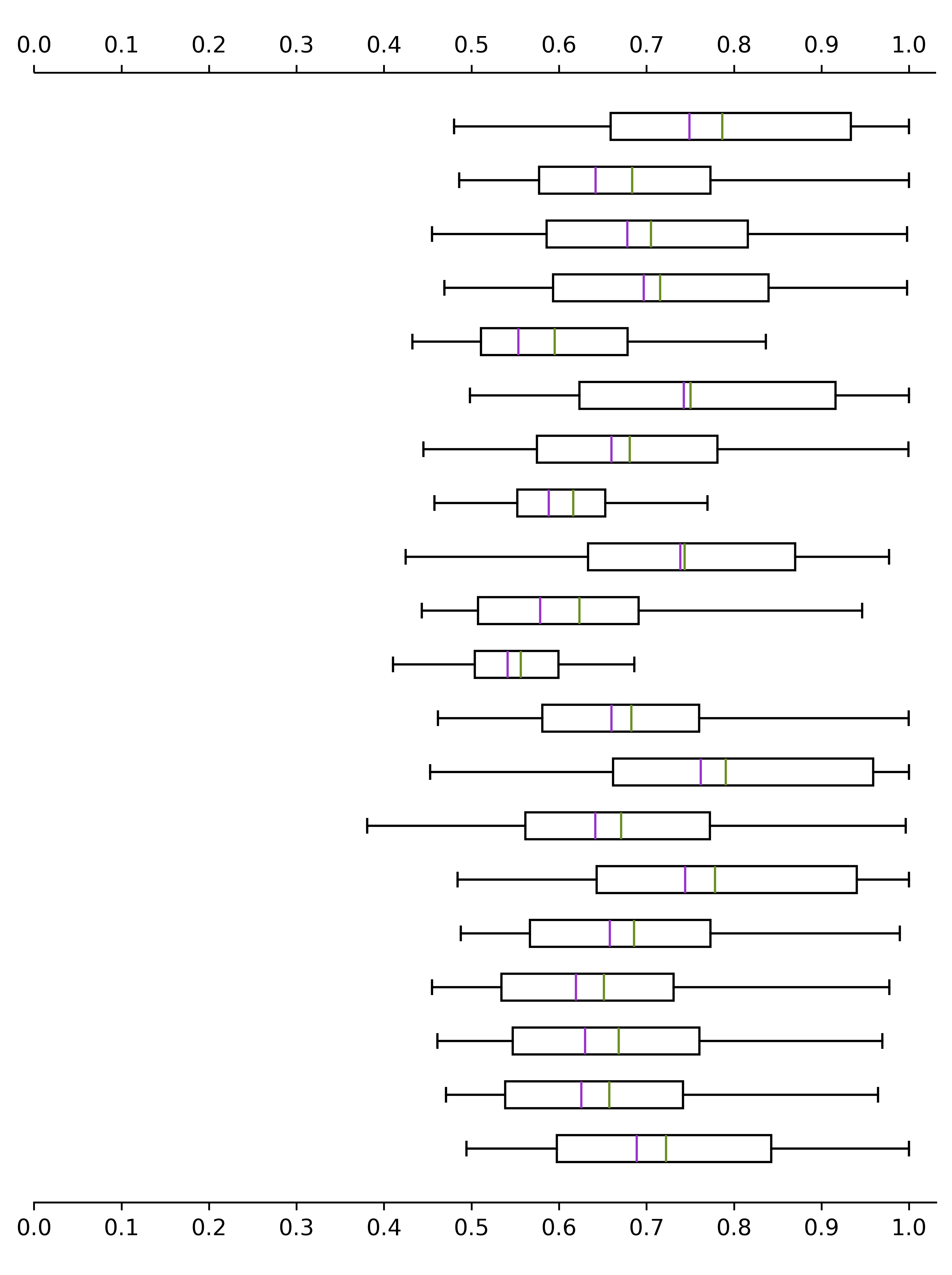}} \\
\hline
\end{tabular}
}
\caption{Distribution of AUC-PR performance for all evaluated algorithms. Deep learning models (blue) and shallow models (yellow) are differentiated by surrogate status (squares for surrogates, triangles for non-surrogates). Mean and median values are shown in green and purple, respectively. Pendant to Table~1 in Section~5.2.}
\label{table:performance_metrics_pr}
\end{table*}

\clearpage
\section{Individual Performance Scores}\label{app:raw}
We state every performance in AUC-ROC in Tables \ref{tab:raw1b}, \ref{tab:raw1}, \ref{tab:raw2b}, \ref{tab:raw2}, \ref{tab:raw3b} and \ref{tab:raw3}. We also give the same performances in AUC-PR in Tables \ref{tab:yraw1b}, \ref{tab:yraw1}, \ref{tab:yraw2b}, \ref{tab:yraw2}, \ref{tab:yraw3b} and \ref{tab:yraw3}.

\begin{table*}
  \caption{AUC-ROC Scores for each datasets and algorithm (1/3|low performing algorithms)}
  \label{tab:raw1b}
\begin{tabular}{lllllllllll}
\hline
 Dataset                & DEAN   & HBOS   & GOAD   & ECOD   & COPOD   & LODA   & NF     & DAGMM   & VAE    & SOD    \\
\hline
 $20news^{2}$           & $56\%$ & $44\%$ & $44\%$ & $44\%$ & $43\%$  & $46\%$ & $49\%$ & $44\%$  & $54\%$ & $46\%$ \\
 $yeast$                & $59\%$ & $42\%$ & $68\%$ & $45\%$ & $38\%$  & $55\%$ & $48\%$ & $49\%$  & $41\%$ & $44\%$ \\
 $vertebral$            & $68\%$ & $38\%$ & $67\%$ & $41\%$ & $34\%$  & $26\%$ & $50\%$ & $50\%$  & $50\%$ & $38\%$ \\
 $MNISTC^{identity}$    & $47\%$ & $49\%$ & $48\%$ & $48\%$ & $48\%$  & $48\%$ & $47\%$ & $49\%$  & $50\%$ & $46\%$ \\
 $speech$               & $59\%$ & $49\%$ & $50\%$ & $48\%$ & $50\%$  & $50\%$ & $47\%$ & $57\%$  & $49\%$ & $35\%$ \\
 $imdb$                 & $53\%$ & $51\%$ & $54\%$ & $48\%$ & $53\%$  & $42\%$ & $53\%$ & $46\%$  & $42\%$ & $46\%$ \\
 $20news^{5}$           & $56\%$ & $49\%$ & $48\%$ & $48\%$ & $47\%$  & $55\%$ & $52\%$ & $48\%$  & $53\%$ & $44\%$ \\
\hline
\end{tabular}

\end{table*}

\begin{table*}
  \caption{AUC-ROC Scores for each datasets and algorithm (1/3|high performing algorithms)}
  \label{tab:raw1}

\begin{tabular}{llllllllllll}
\hline
 Dataset                & DEAN   & LOF    & KNN    & CBLOF   & NeuTral   & AE     & IFor   & PCA    & D.SVDD   & OCSVM   & DTE    \\
\hline
 $20news^{2}$           & $56\%$ & $50\%$ & $48\%$ & $48\%$  & $57\%$    & $44\%$ & $46\%$ & $43\%$ & $44\%$   & $46\%$  & $49\%$ \\
 $yeast$                & $59\%$ & $47\%$ & $45\%$ & $47\%$  & $57\%$    & $42\%$ & $42\%$ & $43\%$ & $45\%$   & $45\%$  & $47\%$ \\
 $vertebral$            & $68\%$ & $53\%$ & $41\%$ & $46\%$  & $59\%$    & $66\%$ & $43\%$ & $41\%$ & $62\%$   & $41\%$  & $39\%$ \\
 $MNISTC^{identity}$    & $47\%$ & $49\%$ & $48\%$ & $50\%$  & $49\%$    & $50\%$ & $48\%$ & $48\%$ & $50\%$   & $47\%$  & $49\%$ \\
 $speech$               & $59\%$ & $53\%$ & $51\%$ & $49\%$  & $44\%$    & $49\%$ & $52\%$ & $49\%$ & $49\%$   & $48\%$  & $56\%$ \\
 $imdb$                 & $53\%$ & $54\%$ & $52\%$ & $53\%$  & $48\%$    & $52\%$ & $49\%$ & $49\%$ & $48\%$   & $48\%$  & $56\%$ \\
 $20news^{5}$           & $56\%$ & $52\%$ & $50\%$ & $55\%$  & $57\%$    & $49\%$ & $46\%$ & $48\%$ & $51\%$   & $49\%$  & $55\%$ \\
\hline
\end{tabular}

\end{table*}

\begin{table*}
  \caption{AUC-ROC Scores for each datasets and algorithm (2/3|low performing algorithms)}
  \label{tab:raw2b}
  
\begin{tabular}{lllllllllll}
\hline
 Dataset                 & DEAN    & HBOS    & GOAD   & ECOD   & COPOD   & LODA   & NF     & DAGMM   & VAE    & SOD    \\
\hline
 $celeba$                & $68\%$  & $77\%$  & $64\%$ & $76\%$ & $75\%$  & $58\%$ & $80\%$ & $62\%$  & $69\%$ & $44\%$ \\
 $CIFAR10^{9}$           & $77\%$  & $60\%$  & $76\%$ & $64\%$ & $61\%$  & $65\%$ & $48\%$ & $62\%$  & $62\%$ & $59\%$ \\
 $FashionMNIST^{6}$      & $82\%$  & $52\%$  & $68\%$ & $60\%$ & $55\%$  & $68\%$ & $55\%$ & $62\%$  & $66\%$ & $44\%$ \\
 $Waveform$              & $73\%$  & $69\%$  & $79\%$ & $58\%$ & $73\%$  & $69\%$ & $67\%$ & $55\%$  & $30\%$ & $49\%$ \\
 $optdigits$             & $99\%$  & $88\%$  & $52\%$ & $52\%$ & $60\%$  & $50\%$ & $55\%$ & $46\%$  & $44\%$ & $21\%$ \\
 $MNISTC^{scale}$        & $89\%$  & $59\%$  & $56\%$ & $59\%$ & $57\%$  & $80\%$ & $53\%$ & $48\%$  & $61\%$ & $17\%$ \\
 $MVTecAD^{cable}$       & $67\%$  & $72\%$  & $66\%$ & $71\%$ & $71\%$  & $71\%$ & $51\%$ & $50\%$  & $51\%$ & $69\%$ \\
\hline
\end{tabular}

\end{table*}

\begin{table*}
  \caption{AUC-ROC Scores for each datasets and algorithm (2/3|high performing algorithms)}
  \label{tab:raw2}
  
\begin{tabular}{llllllllllll}
\hline
 Dataset                 & DEAN    & LOF     & KNN     & CBLOF   & NeuTral   & AE      & IFor   & PCA     & D.SVDD   & OCSVM   & DTE    \\
\hline
 $celeba$                & $68\%$  & $46\%$  & $62\%$  & $59\%$  & $48\%$    & $67\%$  & $69\%$ & $80\%$  & $78\%$   & $72\%$  & $84\%$ \\
 $CIFAR10^{9}$           & $77\%$  & $78\%$  & $71\%$  & $71\%$  & $74\%$    & $73\%$  & $65\%$ & $70\%$  & $62\%$   & $69\%$  & $75\%$ \\
 $FashionMNIST^{6}$      & $82\%$  & $82\%$  & $81\%$  & $74\%$  & $75\%$    & $78\%$  & $63\%$ & $71\%$  & $72\%$   & $65\%$  & $77\%$ \\
 $Waveform$              & $73\%$  & $80\%$  & $81\%$  & $83\%$  & $67\%$    & $70\%$  & $68\%$ & $64\%$  & $68\%$   & $84\%$  & $66\%$ \\
 $optdigits$             & $99\%$  & $100\%$ & $100\%$ & $89\%$  & $64\%$    & $98\%$  & $86\%$ & $52\%$  & $47\%$   & $100\%$ & $50\%$ \\
 $MNISTC^{scale}$        & $89\%$  & $94\%$  & $91\%$  & $84\%$  & $80\%$    & $83\%$  & $66\%$ & $73\%$  & $82\%$   & $65\%$  & $68\%$ \\
 $MVTecAD^{cable}$       & $67\%$  & $78\%$  & $81\%$  & $75\%$  & $71\%$    & $72\%$  & $72\%$ & $71\%$  & $62\%$   & $74\%$  & $72\%$ \\
\hline
\end{tabular}

\end{table*}

\begin{table*}
  \caption{AUC-ROC Scores for each datasets and algorithm (3/3|low performing algorithms)}
  \label{tab:raw3b}
  
\begin{tabular}{lllllllllll}
\hline
 Dataset                   & DEAN    & HBOS    & GOAD    & ECOD    & COPOD   & LODA    & NF     & DAGMM   & VAE     & SOD    \\
\hline
 $shuttle$                 & $100\%$ & $98\%$  & $82\%$  & $99\%$  & $99\%$  & $82\%$  & $9\%$  & $95\%$  & $50\%$  & $31\%$ \\
 $pendigits$               & $99\%$  & $94\%$  & $90\%$  & $92\%$  & $90\%$  & $88\%$  & $67\%$ & $64\%$  & $89\%$  & $21\%$ \\
 $MNISTC^{glass\_blur}$    & $100\%$ & $90\%$  & $92\%$  & $89\%$  & $88\%$  & $90\%$  & $74\%$ & $62\%$  & $52\%$  & $54\%$ \\
 $cardio$                  & $89\%$  & $84\%$  & $95\%$  & $93\%$  & $91\%$  & $95\%$  & $90\%$ & $69\%$  & $95\%$  & $45\%$ \\
 $http$                    & $100\%$ & $99\%$  & $1\%$   & $97\%$  & $99\%$  & $43\%$  & $99\%$ & $99\%$  & $100\%$ & $40\%$ \\
 $MVTecAD^{bottle}$        & $96\%$  & $96\%$  & $92\%$  & $92\%$  & $96\%$  & $95\%$  & $91\%$ & $50\%$  & $7\%$   & $97\%$ \\
 $Stamps$                  & $89\%$  & $90\%$  & $88\%$  & $90\%$  & $91\%$  & $91\%$  & $89\%$ & $72\%$  & $91\%$  & $52\%$ \\
\hline
\end{tabular}

\end{table*}

\begin{table*}
  \caption{AUC-ROC Scores for each datasets and algorithm (3/3|high performing algorithms)}
  \label{tab:raw3}
  \begin{tabular}{llllllllllll}
\hline
 Dataset                   & DEAN    & LOF     & KNN     & CBLOF   & NeuTral   & AE      & IFor    & PCA     & D.SVDD   & OCSVM   & DTE     \\
\hline
 $shuttle$                 & $100\%$ & $100\%$ & $100\%$ & $99\%$  & $100\%$   & $100\%$ & $100\%$ & $99\%$  & $99\%$   & $100\%$ & $50\%$  \\
 $pendigits$               & $99\%$  & $99\%$  & $100\%$ & $97\%$  & $62\%$    & $89\%$  & $98\%$  & $93\%$  & $94\%$   & $94\%$  & $98\%$  \\
 $MNISTC^{glass\_blur}$    & $100\%$ & $99\%$  & $100\%$ & $98\%$  & $96\%$    & $99\%$  & $95\%$  & $96\%$  & $97\%$   & $92\%$  & $99\%$  \\
 $cardio$                  & $89\%$  & $93\%$  & $91\%$  & $92\%$  & $86\%$    & $91\%$  & $93\%$  & $95\%$  & $92\%$   & $94\%$  & $92\%$  \\
 $http$                    & $100\%$ & $93\%$  & $100\%$ & $99\%$  & $100\%$   & $99\%$  & $99\%$  & $100\%$ & $100\%$  & $100\%$ & $99\%$  \\
 $MVTecAD^{bottle}$        & $96\%$  & $96\%$  & $96\%$  & $97\%$  & $96\%$    & $96\%$  & $97\%$  & $96\%$  & $96\%$   & $96\%$  & $95\%$  \\
 $Stamps$                  & $89\%$  & $93\%$  & $95\%$  & $93\%$  & $99\%$    & $90\%$  & $92\%$  & $92\%$  & $93\%$   & $91\%$  & $92\%$  \\
\hline
\end{tabular}

\end{table*}

\begin{table*}
  \caption{AUC-PR Scores for each datasets and algorithm (1/3|low performing algorithms)}
  \label{tab:yraw1b}
  \begin{tabular}{lllllllllll}
\hline
 Dataset              & DEAN   & GOAD   & HBOS   & ECOD   & COPOD   & LODA   & NF     & VAE    & DAGMM   & SOD    \\
\hline
 $20news^{2}$         & $56\%$ & $45\%$ & $44\%$ & $45\%$ & $44\%$  & $44\%$ & $44\%$ & $52\%$ & $45\%$  & $45\%$ \\
 $imdb$               & $53\%$ & $50\%$ & $48\%$ & $46\%$ & $49\%$  & $48\%$ & $54\%$ & $44\%$ & $46\%$  & $46\%$ \\
 $WPBC$               & $54\%$ & $46\%$ & $48\%$ & $47\%$ & $49\%$  & $50\%$ & $49\%$ & $49\%$ & $50\%$  & $45\%$ \\
 $MNISTC^{identity}$  & $47\%$ & $49\%$ & $49\%$ & $49\%$ & $49\%$  & $49\%$ & $48\%$ & $50\%$ & $50\%$  & $48\%$ \\
 $vertebral$          & $68\%$ & $58\%$ & $42\%$ & $43\%$ & $39\%$  & $37\%$ & $75\%$ & $50\%$ & $50\%$  & $41\%$ \\
 $yeast$              & $59\%$ & $61\%$ & $49\%$ & $50\%$ & $46\%$  & $54\%$ & $49\%$ & $45\%$ & $53\%$  & $45\%$ \\
 $20news^{1}$         & $64\%$ & $50\%$ & $49\%$ & $47\%$ & $49\%$  & $45\%$ & $47\%$ & $45\%$ & $47\%$  & $50\%$ \\
\hline
\end{tabular}

\end{table*}

\begin{table*}
  \caption{AUC-PR Scores for each datasets and algorithm (1/3|high performing algorithms)}
  \label{tab:yraw1}

\begin{tabular}{llllllllllll}
\hline
 Dataset              & DEAN   & LOF    & KNN    & NeuTral   & CBLOF   & PCA    & DTE    & IFor   & OCSVM   & D.SVDD   & AE     \\
\hline
 $20news^{2}$         & $56\%$ & $49\%$ & $47\%$ & $57\%$    & $44\%$  & $44\%$ & $51\%$ & $44\%$ & $45\%$  & $43\%$   & $47\%$ \\
 $imdb$               & $53\%$ & $52\%$ & $48\%$ & $50\%$    & $49\%$  & $47\%$ & $53\%$ & $49\%$ & $46\%$  & $45\%$   & $49\%$ \\
 $WPBC$               & $54\%$ & $50\%$ & $50\%$ & $46\%$    & $47\%$  & $51\%$ & $48\%$ & $52\%$ & $52\%$  & $51\%$   & $48\%$ \\
 $MNISTC^{identity}$  & $47\%$ & $50\%$ & $49\%$ & $50\%$    & $51\%$  & $49\%$ & $50\%$ & $50\%$ & $48\%$  & $49\%$   & $54\%$ \\
 $vertebral$          & $68\%$ & $51\%$ & $43\%$ & $60\%$    & $47\%$  & $42\%$ & $40\%$ & $42\%$ & $53\%$  & $44\%$   & $68\%$ \\
 $yeast$              & $59\%$ & $50\%$ & $49\%$ & $57\%$    & $48\%$  & $47\%$ & $49\%$ & $47\%$ & $48\%$  & $45\%$   & $48\%$ \\
 $20news^{1}$         & $64\%$ & $60\%$ & $61\%$ & $63\%$    & $50\%$  & $48\%$ & $48\%$ & $49\%$ & $48\%$  & $52\%$   & $53\%$ \\
\hline
\end{tabular}

\end{table*}

\begin{table*}
  \caption{AUC-PR Scores for each datasets and algorithm (2/3|low performing algorithms)}
  \label{tab:yraw2b}
  
\begin{tabular}{lllllllllll}
\hline
 Dataset                 & DEAN    & GOAD   & HBOS   & ECOD   & COPOD   & LODA   & NF     & VAE     & DAGMM   & SOD    \\
\hline
 $MVTecAD^{metal\_nut}$  & $67\%$  & $74\%$ & $60\%$ & $62\%$ & $60\%$  & $71\%$ & $56\%$ & $48\%$  & $50\%$  & $68\%$ \\
 $SpamBase$              & $68\%$  & $54\%$ & $76\%$ & $61\%$ & $63\%$  & $72\%$ & $77\%$ & $50\%$  & $55\%$  & $53\%$ \\
 $celeba$                & $68\%$  & $66\%$ & $78\%$ & $77\%$ & $76\%$  & $58\%$ & $73\%$ & $69\%$  & $59\%$  & $42\%$ \\
 $optdigits$             & $99\%$  & $48\%$ & $84\%$ & $47\%$ & $52\%$  & $46\%$ & $64\%$ & $46\%$  & $45\%$  & $35\%$ \\
 $CIFAR10^{9}$           & $77\%$  & $75\%$ & $62\%$ & $65\%$ & $63\%$  & $71\%$ & $53\%$ & $64\%$  & $63\%$  & $59\%$ \\
 $CIFAR10^{8}$           & $74\%$  & $78\%$ & $63\%$ & $65\%$ & $64\%$  & $75\%$ & $56\%$ & $65\%$  & $62\%$  & $57\%$ \\
 $CIFAR10^{6}$           & $77\%$  & $72\%$ & $65\%$ & $66\%$ & $65\%$  & $62\%$ & $60\%$ & $61\%$  & $56\%$  & $61\%$ \\
\hline
\end{tabular}

\end{table*}

\begin{table*}
  \caption{AUC-PR Scores for each datasets and algorithm (2/3|high performing algorithms)}
  \label{tab:yraw2}
  
\begin{tabular}{llllllllllll}
\hline
 Dataset                 & DEAN    & LOF     & KNN     & NeuTral   & CBLOF   & PCA    & DTE    & IFor   & OCSVM   & D.SVDD   & AE      \\
\hline
 $MVTecAD^{metal\_nut}$  & $67\%$  & $77\%$  & $78\%$  & $72\%$    & $77\%$  & $68\%$ & $80\%$ & $68\%$ & $70\%$  & $71\%$   & $59\%$  \\
 $SpamBase$              & $68\%$  & $64\%$  & $75\%$  & $45\%$    & $70\%$  & $79\%$ & $66\%$ & $80\%$ & $76\%$  & $79\%$   & $72\%$  \\
 $celeba$                & $68\%$  & $45\%$  & $62\%$  & $49\%$    & $67\%$  & $81\%$ & $77\%$ & $74\%$ & $73\%$  & $76\%$   & $67\%$  \\
 $optdigits$             & $99\%$  & $100\%$ & $100\%$ & $66\%$    & $72\%$  & $46\%$ & $75\%$ & $79\%$ & $100\%$ & $43\%$   & $97\%$  \\
 $CIFAR10^{9}$           & $77\%$  & $78\%$  & $72\%$  & $75\%$    & $70\%$  & $70\%$ & $73\%$ & $67\%$ & $69\%$  & $68\%$   & $63\%$  \\
 $CIFAR10^{8}$           & $74\%$  & $76\%$  & $70\%$  & $74\%$    & $69\%$  & $70\%$ & $68\%$ & $66\%$ & $72\%$  & $66\%$   & $66\%$  \\
 $CIFAR10^{6}$           & $77\%$  & $76\%$  & $75\%$  & $75\%$    & $71\%$  & $70\%$ & $75\%$ & $67\%$ & $66\%$  & $72\%$   & $68\%$  \\
\hline
\end{tabular}

\end{table*}

\begin{table*}
  \caption{AUC-PR Scores for each datasets and algorithm (3/3|low performing algorithms)}
  \label{tab:yraw3b}
  
\begin{tabular}{lllllllllll}
\hline
 Dataset                   & DEAN    & GOAD    & HBOS    & ECOD    & COPOD   & LODA    & NF     & VAE     & DAGMM   & SOD    \\
\hline
 $musk$                    & $53\%$  & $74\%$  & $100\%$ & $97\%$  & $95\%$  & $99\%$  & $76\%$ & $50\%$  & $92\%$  & $31\%$ \\
 $pendigits$               & $99\%$  & $84\%$  & $92\%$  & $90\%$  & $87\%$  & $96\%$  & $58\%$ & $88\%$  & $56\%$  & $37\%$ \\
 $smtp$                    & $92\%$  & $89\%$  & $88\%$  & $91\%$  & $93\%$  & $92\%$  & $96\%$ & $36\%$  & $89\%$  & $65\%$ \\
 $MNISTC^{glass\_blur}$    & $100\%$ & $89\%$  & $86\%$  & $84\%$  & $83\%$  & $93\%$  & $79\%$ & $49\%$  & $63\%$  & $52\%$ \\
 $cardio$                  & $89\%$  & $94\%$  & $85\%$  & $90\%$  & $89\%$  & $89\%$  & $91\%$ & $91\%$  & $69\%$  & $48\%$ \\
 $shuttle$                 & $100\%$ & $87\%$  & $99\%$  & $99\%$  & $100\%$ & $90\%$  & $32\%$ & $50\%$  & $95\%$  & $42\%$ \\
 $WBC$                     & $99\%$  & $99\%$  & $99\%$  & $100\%$ & $100\%$ & $94\%$  & $71\%$ & $99\%$  & $50\%$  & $70\%$ \\
\hline
\end{tabular}

\end{table*}

\begin{table*}
  \caption{AUC-PR Scores for each datasets and algorithm (3/3|high performing algorithms)}
  \label{tab:yraw3}
  
\begin{tabular}{llllllllllll}
\hline
 Dataset                   & DEAN    & LOF     & KNN     & NeuTral   & CBLOF   & PCA     & DTE     & IFor    & OCSVM   & D.SVDD   & AE      \\
\hline
 $musk$                    & $53\%$  & $100\%$ & $100\%$ & $99\%$    & $100\%$ & $100\%$ & $43\%$  & $96\%$  & $75\%$  & $100\%$  & $100\%$ \\
 $pendigits$               & $99\%$  & $98\%$  & $100\%$ & $61\%$    & $98\%$  & $90\%$  & $98\%$  & $96\%$  & $91\%$  & $78\%$   & $84\%$  \\
 $smtp$                    & $92\%$  & $95\%$  & $95\%$  & $77\%$    & $90\%$  & $89\%$  & $92\%$  & $80\%$  & $88\%$  & $87\%$   & $65\%$  \\
 $MNISTC^{glass\_blur}$    & $100\%$ & $99\%$  & $100\%$ & $94\%$    & $97\%$  & $95\%$  & $99\%$  & $91\%$  & $89\%$  & $95\%$   & $93\%$  \\
 $cardio$                  & $89\%$  & $91\%$  & $90\%$  & $85\%$    & $89\%$  & $94\%$  & $90\%$  & $93\%$  & $91\%$  & $97\%$   & $91\%$  \\
 $shuttle$                 & $100\%$ & $100\%$ & $99\%$  & $98\%$    & $98\%$  & $99\%$  & $75\%$  & $100\%$ & $100\%$ & $99\%$   & $99\%$  \\
 $WBC$                     & $99\%$  & $92\%$  & $99\%$  & $73\%$    & $99\%$  & $99\%$  & $45\%$  & $99\%$  & $99\%$  & $97\%$   & $98\%$  \\
\hline
\end{tabular}

\end{table*}

\end{document}